\documentclass[a4paper,sn-mathphys,Numbered]{sn-jnl}

\usepackage{graphicx}%
\usepackage{multirow}%
\usepackage{amsmath,amssymb,amsfonts}%
\usepackage{amsthm}%
\usepackage{mathrsfs}%
\usepackage[title]{appendix}%
\usepackage{xcolor}%
\usepackage{textcomp}%
\usepackage{manyfoot}%
\usepackage{booktabs}%
\usepackage{algorithm}%
\usepackage{algorithmicx}%
\usepackage{algpseudocode}%
\usepackage{listings}%
\usepackage{tabularx}
\usepackage[figuresright]{rotating}
\usepackage{array}
\usepackage{adjustbox}
\usepackage{longtable} 
\usepackage{lscape}   
\usepackage{chngcntr}
\counterwithout{figure}{section}
\renewcommand{\thefigure}{\arabic{figure}}
\renewcommand{\thetable}{\arabic{table}}
\usepackage{threeparttable}

\theoremstyle{thmstyleone}%
\theoremstyle{thmstyletwo}%

\theoremstyle{thmstylethree}%

\begin{document}

\title[Learning Interpretable Network Dynamics via Universal Neural Symbolic Regression]{Learning Interpretable Network Dynamics via Universal Neural Symbolic Regression}


\author[1,2]{\fnm{Jiao} \sur{Hu}}

\author*[1,2]{\fnm{Jiaxu} \sur{Cui}}\email{cjx@jlu.edu.cn}



\author*[1,2]{\fnm{Bo} \sur{Yang}}\email{ybo@jlu.edu.cn}

\affil[1]{\orgdiv{College of Computer Science and Technology}, \orgname{Jilin University}, \orgaddress{ \city{Changchun}, \postcode{130012}, \country{China}}}

\affil[2]{\orgdiv{Key Laboratory of Symbolic Computation and Knowledge Engineering of Ministry of Education}, \orgname{Jilin University}, \orgaddress{\city{Changchun}, \postcode{130012}, \country{China}}}



\abstract{
Discovering governing equations of complex network dynamics is a fundamental challenge in contemporary science with rich data, which can uncover the mysterious patterns and mechanisms of the formation and evolution of complex phenomena in various fields and assist in decision-making.
In this work, we develop a universal computational tool that can automatically, efficiently, and accurately learn the symbolic changing patterns of complex system states by combining the excellent fitting ability from deep learning and the equation inference ability from pre-trained symbolic regression.
{We conduct intensive experimental verifications on more than ten representative scenarios from physics, biochemistry, ecology, epidemiology, etc.  
Results demonstrate the outstanding effectiveness and efficiency of our tool by comparing with the state-of-the-art symbolic regression techniques for network dynamics.
The application to real-world systems including global epidemic transmission and pedestrian movements has verified its practical applicability.}
We believe that our tool can serve as a universal solution to dispel the fog of hidden mechanisms of changes in complex phenomena, advance toward interpretability, and inspire more scientific discoveries.
}




\maketitle

\section{Introduction}\label{sec1}

From the Book of Changes in ancient China to the dialectical thinking in the West, there exists a common philosophical thought that \textit{the only constant is change}.
Undoubtedly, scientists have been striving to discover the laws of changes in complex phenomena, attempting to explain, forecast, and regulate all things \cite{tyutyunnikdisorder}, {such as emergence \cite{delvenne2015diffusion}, chaos \cite{sprott2008chaotic}, synchronization \cite{rodrigues2016kuramoto}, and critical phenomena \cite{bernaschi2024quantum}}.
As a widely accepted modeling, the changing patterns of states from complex systems are generally governed by a set of nonlinear differential equations \cite{barzel2013universality} as
$
\dot{X}(t)=f(X(t),A,t),
$
where $X(t)\in\mathbb{R}^{N\times d}$ is the system states at time $t$, 
$N$ and $d$ are the number of system components (nodes) and the state dimension, respectively.
$A$ represents the extra information beyond the system states, such as topological interactions among system components.
As shown in the above formula, the dynamic behaviors exhibited by complex systems are primarily contingent upon the intricate interdependence between their internal interactions $A$ and dynamics governing equations $f$ \cite{barzel2013universality, harush2017dynamic}.
This prompts people to seek reliable methodologies to formulate dynamics models of these complex systems \cite{gao2022autonomous,ShiDCYL023}.
However, a remarkable challenge arises in this pursuit.
In theoretically complete physical systems, laws of changes are delineated by well-discovered foundational principles \cite{cranmer2020discovering, udrescu2020ai, zhang2022physics}, such as the electromagnetic laws dictating the microscale exchanges among propelled particles.
For the majority of complex systems, $f$ is agnostic, and equivalent foundational rules remain incompletely elucidated, such as global epidemic outbreak \cite{murphy2021deep}, extreme climate anomalies \cite{zhang2023skilful}, and extinction of biological populations \cite{gao2016universal}.
Consequently, this vague development has limited the exploration of these complex fields.

Fortunately, in the current era of data acquisition gradually becoming easier, the emergence of data-driven technologies has assisted in increasing the frequency with which human experts discover system change patterns \cite{timme2007revealing, napoletani2008reconstructing, maslov2007propagation,pauleve2020reconciling,ding2024artificial}.
They can provide domain experts with richer and more meaningful inspiration for various fields, accelerating the process of scientific discovery, such as mathematics \cite{NatureAIMath,LyapunovNIPS2024} and physics \cite{huang2021coupled, bottcher2022ai}.
Although much excellent work has been developed to reconstruct the symbolic models for low-variate dynamics of complex systems \cite{SRReview2024}, 
{e.g., bivariate shear flow equation \cite{cava2021contemporary}, trivariate metapopulation epidemic model \cite{Lipshtatmetapopulation}, and up to 9-variate Newton's law of gravitation \cite{udrescu2020ai}},
inferring governing equations for high-variate network dynamics remains important and challenging.
This is mainly because the number of nodes $N$ in network dynamics is usually large, such as the epidemic spreading with transmission areas or individual numbers ranging from tens to billions \cite{murphy2021deep}, and $d$ is sometimes multi-dimensional, resulting in too many free variables ($N\times d$) in the equations and topological interactions with exponential growth, thereby increasing the complexity of inferring symbolic models \cite{gao2016universal}.

At present, several cutting-edge work is attempting to deal with the discovery of governing equations from network dynamics \cite{gao2023data,ding2024artificial}.
{Two-phase sparse symbolic regression (TPSINDy) \cite{gao2022autonomous} simply parameterizes $f$ as a learnable linear combination of 
pre-defined orthogonal or non-orthogonal elementary function terms. 
Although its equation inference efficiency is high, the rationality of pre-defined function terms directly affects the inference results, so sufficient and correct domain expert knowledge is usually required \cite{chen2021physics}.
Another group of methods of using graph neural networks (GNN) to parameterize $f$ overcomes excessive expert knowledge \cite{zang2020neural}. Still, due to the use of genetic programming (GP) to parse neural networks into symbolic equations, it brings the high-cost evolutionary search efficiency issue \cite{cranmer2020discovering,zhang2023skilful}.}
Therefore, how to effectively balance expert knowledge and computational costs, while ensuring high computational efficiency, introducing only a small amount or no expert knowledge, lowering the threshold for use, and efficiently discovering governing equations remains a gap.

To address the challenges above, we develop a universal neural symbolic regression tool that can automatically, efficiently, and accurately learn the changing patterns of complex system states by combining the excellent fitting ability from deep learning and the equation inference ability from pre-trained symbolic regression.
Our analysis of various complex network dynamics scenarios from physics, biochemistry, ecology, epidemiology, etc., 
indicates that our tool has outstanding effectiveness and efficiency.
It can accurately and efficiently discover the governing equations of network dynamics, even in the face of noisy and topologically missing data, and has achieved excellent results in chaotic systems and real-world systems including global epidemic transmission and pedestrian movements.
We believe that our tool can serve as a new and general solution to eliminate the fog of hidden mechanisms of changes in complex phenomena from broad fields.

\section{Results}\label{sec2}

In this work, we wonder to develop a computational tool LLC (\underline{L}earning \underline{L}aw of \underline{C}hanges) to discover the ordinary differential equations (ODEs) from observed data of network dynamics, i.e., 
$
\text{LLC}: \mathcal{O}\to \mathcal{F}
$,
where observations $O=\{(X(t),A,M_x,M_a)\}_{t=0}^{T}\subseteq\mathcal{O}$ and the ODEs $f\in\mathcal{F}$.
In fact, observations are sampled from continuous state data, which is obtained by solving the initial value problem with given initial states $X(0)$ and topology $A$, i.e., $X(t)=X(0)+\int_0^{t}{f(X(\tau),A,\tau)}d\tau$.
Note that $M_x$ and $M_a$ in observations $O$ are the masks for observed states and topological structure respectively, with the same shapes of $X(t)$ and $A$, depicting the incompleteness of the observations.
By configuring different masks, it can flexibly handle complex scenarios, such as heterogeneous network dynamics with local observations.
The following introduces the overall process of our tool and the analysis of experimental results on sufficient network dynamics scenarios.

\begin{figure}[t]
    \centering   
    \includegraphics[width=\textwidth]{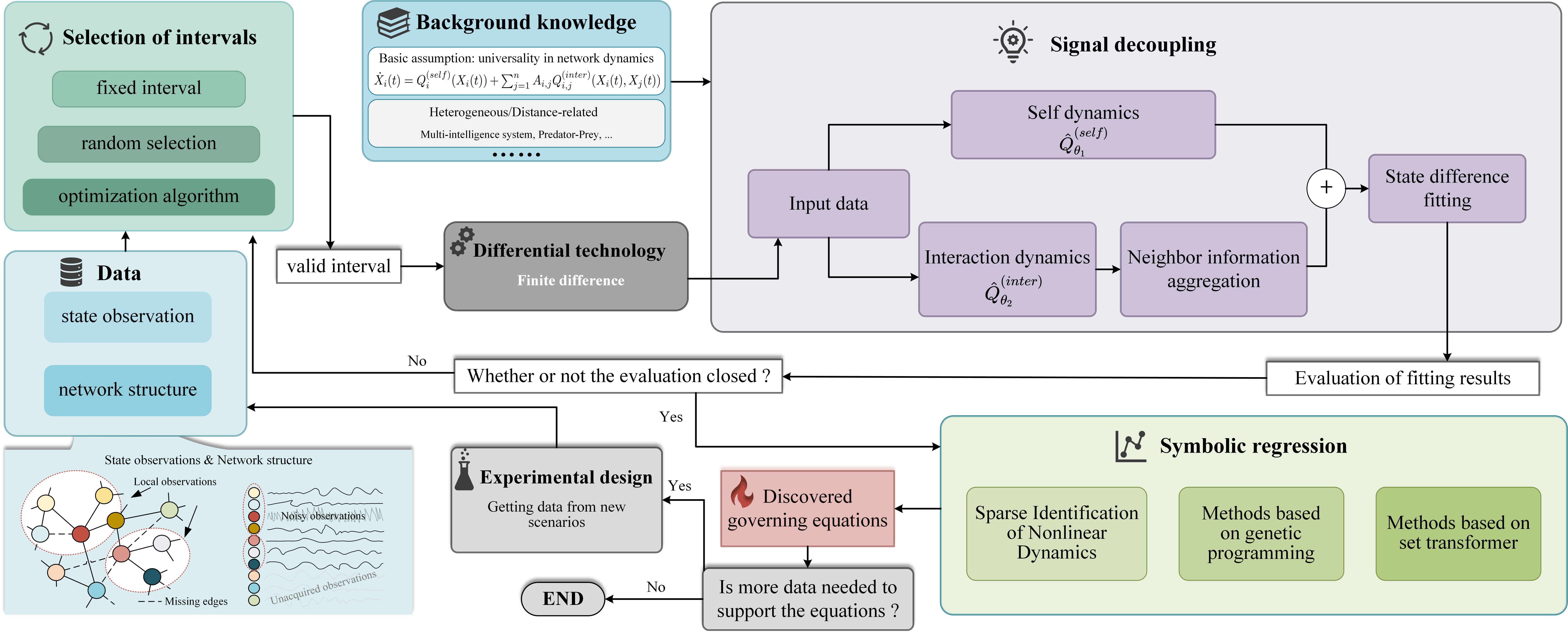}    
    \caption{
    The overall process of the LLC (Learning Law of Changes).
    Observed data can be acquired from the initial experiments on the new scenario, including system states over time and topology, i.e. $O$.
    An interval selection strategy is to choose valid interval data and then we can get differentials, i.e. $\dot{X}_i(t)$, through finite difference on ${X}_i(t)$.
    Combined with physical priors, the neural networks, i.e., $\Hat{Q}_{\theta_1}^{(self)}$ and $\Hat{Q}_{\theta_2}^{(inter)}$, are to decouple the network dynamics signals and achieve variable reduction, until learning to meet the fitting requirements.
    Otherwise, the training is repeated.
    After having well-fitted neural networks, we use symbolic regression techniques to parse their approximate white-box equations efficiently.
    Of course, suppose more observed data is needed to support the discovery. In that case, the experimental design can be revisited until the satisfactory governing equations of network dynamics are finally acquired to break out of the loop.
    }
    \label{fig1}
\end{figure}

\subsection{The overall process of the LLC (Learning Law of Changes)}


When developing the computational tool, we encounter two stubborn issues: high-dimensional free variables and efficiency issues in symbolic inference.
As mentioned earlier, due to the excessive number of nodes and state dimensions, the combination space of the symbolic models is too huge to seek the well-fitted dynamics equations for observed data \cite{virgolin2022symbolic}.
Meanwhile, if we simply test all symbolic models by increasing equation length, it may take longer than the age of our universe until we get the target \cite{udrescu2020ai}.
Even if de novo optimization algorithms such as evolutionary computation \cite{tenachi2023deep}, reinforcement learning \cite{petersen2020deep}, and Monte Carlo tree search \cite{sun2022symbolic}
are applied to find the objective symbolic model, it will still take an intolerable amount of time.
More importantly, the performance of existing symbolic regression methods drops sharply when the number of variables exceeds three \cite{biggio2021neural}.
Therefore, to tackle both difficulties, we employ the divide and conquer approach by introducing a few physical priors and using the excellent fitting power from neural networks to achieve dimensionality reduction of high-dimensional network dynamic signals, then using a pre-trained symbolic regression model to accelerate the efficiency of inferring dynamics equations.

\textbf{Decoupling network dynamics signals through neural networks and physical priors}.
To alleviate the curse of dimensionality in network dynamics, we introduce a physical prior in network dynamics that the change of network states can be influenced by its own states and the states of its neighbors \cite{barzel2013universality,gao2022autonomous,liu2023we,SCIS_ours}.
That is, we can decompose the governing equation $f$ into two coupled components: self dynamics ($Q^{(self)}$) and interaction dynamics ($Q^{(inter)}$), thereby the governing equations of network dynamics can be reformalized in node-wise form as
$
\dot{X}_i(t)=Q_i^{(self)}(X_i(t))+\sum_{j=1}^{N}{A_{i,j}Q_{i,j}^{(inter)}(X_i(t),X_j(t))}
$,
where the subscripts $i$ and $j$ represent the corresponding nodes, $A_{i,j}$ is the adjacency matrix, $Q_i^{(self)}$ and $Q_{i,j}^{(inter)}$ capture the evolution process of their own states and the neighbors' dynamical mechanism governing the pairwise interactions respectively.
At this moment, $f$ is composed of a set of functions, i.e., $f:=\{Q_i^{(self)},Q_{i,j}^{(inter)}\}_{i,j=1}^{N}$.
Actually, with the appropriate choice of $Q_i^{(self)}$ and $Q_{i,j}^{(inter)}$, the above equation can describe broad network dynamics scenarios \cite{barzel2013universality,barzel2015constructing}.
Unlike the popular message-passing-based graph neural networks \cite{velivckovic2018graph}, the fusion with this physical prior has been empirically validated to learn underlying dynamics and is more suitable for handling network dynamics scenarios \cite{liu2023we}.
Significantly, such formulation can achieve the desired dimensionality reduction for high-dimensional network dynamics, by learning the $d$-variate $Q_i^{(self)}$ and $2d$-variate $Q_{i,j}^{(inter)}$ instead of directly inferring the $(n\times d)$-variate $f$.
Note that if the network dynamics are homogeneous, meaning that the governing equations for all nodes are consistent, then only two functions need to be learned. 
However, if the dynamics are heterogeneous, multiple sets of functions need to be learned according to different settings.
For ease of description, we elaborate on homogeneous scenarios here, i.e., omitting subscripts $i$ and $j$, and the setting for heterogeneous scenarios can be acquired from subsequent experiments.
As we know, the differential signal of network dynamics ($\dot{X}_i(t)$) can be calculated by $Q^{(self)}$ and $Q^{(inter)}$. 
Thanks to the excellent nonlinear fitting power from neural networks, we parameterize them separately using two neural networks, i.e., $\Hat{Q}_{\theta_1}^{(self)}$ and $\Hat{Q}_{\theta_2}^{(inter)}$. 
{More specific architectures can be found in Methods.}
By fitting $\dot{X}_i(t)$, the parameters $\theta_1$ and $\theta_2$ in the neural networks can be learned through backpropagation.
Now, we achieve decoupling of the dynamics signals by obtaining well-trained self dynamics and interaction dynamics functions.




\textbf{Parsing governing equations via pre-trained symbolic regression}.
To open the black-boxes of the trained $\Hat{Q}_{\theta_1}^{(self)}$ and $\Hat{Q}_{\theta_2}^{(inter)}$,
we use symbolic regression techniques \cite{SRReview2024} to find the readable white-box equations that approximate the neural networks best from a vast equation space.
However, the principle behind conventional symbolic search algorithms is to seek optimal symbolic models from scratch when facing new tasks, resulting in typically long execution times \cite{petersen2020deep,xureinforcement,sun2022symbolic,tenachi2023deep}.
As a new generation of symbolic regression techniques, pre-trained transformer models \cite{biggio2021neural} can improve the efficiency of equation inference.
Such models are pre-trained on vast sample data consisting of input-output points from neural networks and equation pairs \cite{lee2019set}, integrating massive equation knowledge, to quickly obtain corresponding equations from input-output points with solving new tasks.
When inferring a new equation, only one forward propagation is required, significantly reducing the inference time, and then the constants in the equations can be optionally fine-tuned by using BFGS \cite{fletcher2000practical} or stochastic gradient optimization \cite{bottou2010large} to ensure inference accuracy.
Specifically, we empirically use the NSRA \cite{biggio2021neural}  as our symbolic regression model, which is pre-trained on a dataset with hundreds of millions of equations and performs the best among several pre-trained symbolic regression models, to efficiently parse the symbolic equations of $\Hat{Q}_{\theta_1}^{(self)}$ and $\Hat{Q}_{\theta_2}^{(inter)}$.
To our knowledge, this is the first attempt to introduce pre-trained models to discover governing equations of network dynamics.


\textbf{Assembling all modules to form an executable pipeline}.
After overcoming the two core issues mentioned above, we built a tool to open up the path for discovering governing equations of network dynamics, as shown in Fig.~\ref{fig1}.
We obtain observed data from the initial experiments on the new scenario, including system states over time and topology, i.e. $O$, and use an interval selection strategy to choose valid interval data for getting differentials, i.e. $\dot{X}_i(t)$, through finite difference on ${X}_i(t)$.
Then, combined with physical priors, the neural networks, i.e., $\Hat{Q}_{\theta_1}^{(self)}$ and $\Hat{Q}_{\theta_2}^{(inter)}$, are to decouple the network dynamics signals and achieve free-variable reduction, until learning to meet the fitting requirements.
Otherwise, the training is repeated.
Although we can directly train the neural networks via the adjoint method of neural ODE \cite{huang2021coupled,zang2020neural,liu2023we} based on observed data (${X}_i(t)$), we have found empirically that fitting the neural networks on differentials can bypass the high computational complexity of numerical integration in neural ODE and have more stable performance.
After having well-fitted $\Hat{Q}_{\theta_1}^{(self)}$ and $\Hat{Q}_{\theta_2}^{(inter)}$, we use symbolic regression techniques to parse their approximate white-box equations efficiently.
Of course, suppose more observed data is needed to support the discovery. In that case, the experimental design can be revisited until the satisfactory governing equations of network dynamics are finally acquired to break out of the loop.
The modules in the pipeline will be detailed in Methods.
Next, we will report and analyze the results of testing the LLC on sufficient network dynamics from various fields.

\begin{figure}[t]
    \centering    
    \includegraphics[width=\textwidth]{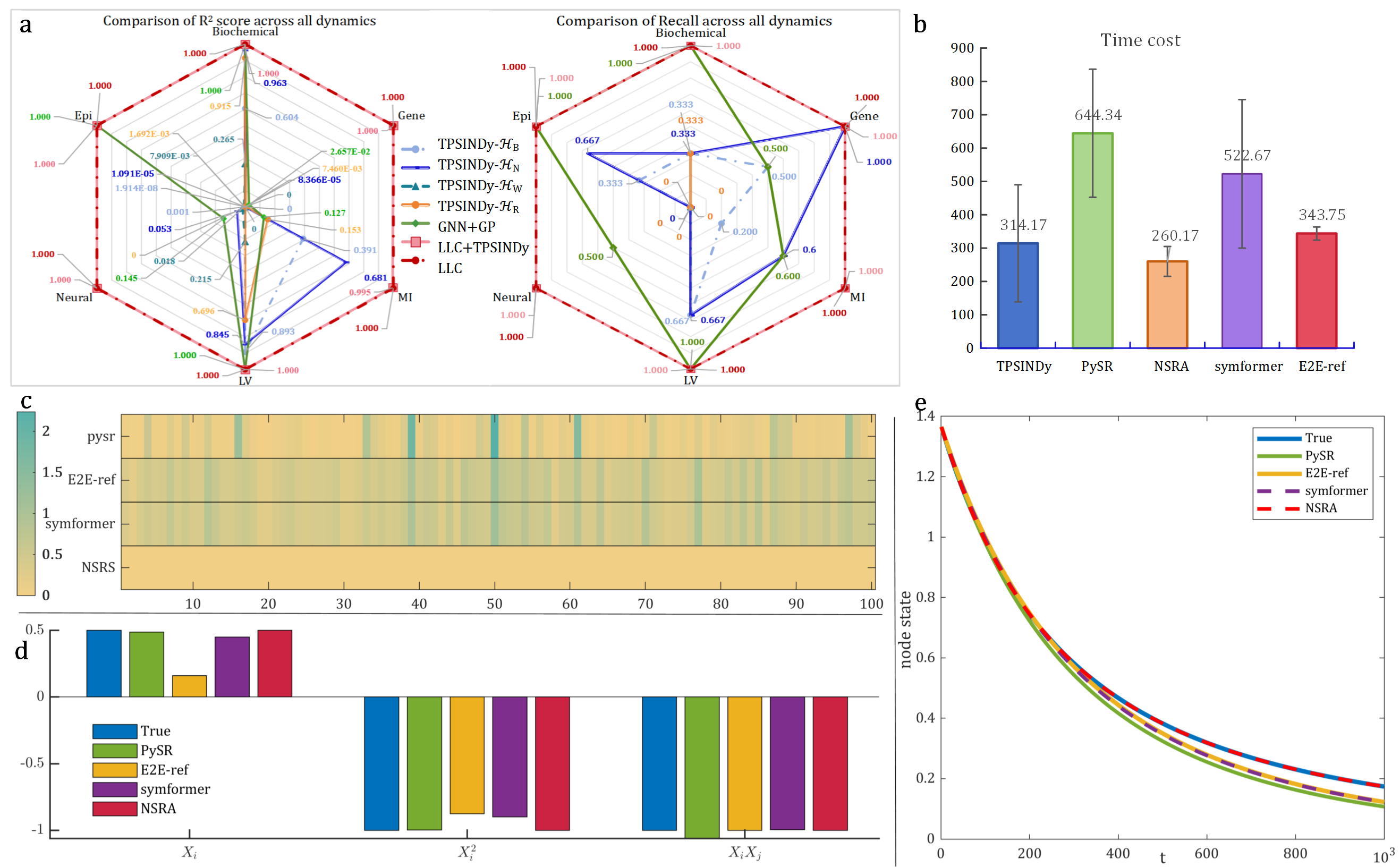}    
    \caption{
    {Results of inferring one-dimensional homogeneous network dynamics. 
    \textbf{a.} Comparison of the accuracy on predictions ($R^2$ score) and discovered equations (Recall) for reconstructing dynamics from six scenarios, including Biochemical (Bio),  Gene regulatory (Gene), Mutualistic Interaction (MI), Lotka-Volterra  (LV), Neural (Neur), and Epidemic (Epi) dynamics.
    TPSINDy's results are highly dominated by its choice of function terms and our LLC significantly outperforms the comparative methods covering all network dynamics scenarios. 
    \textbf{b.} Comparison of the average execution times across all dynamics for various symbolic regression methods.
    \textbf{c.} The NED (Normalized Estimation Error) between the predictive results produced by the discovered governing equations and ground truth in the LV scenario.
    \textbf{d.} Comparison of fitting coefficients in governing equations discovered by symbolic regression methods.
    \textbf{e.} Comparison of state prediction curves for an individual node.}
    }
    \label{fig2}
\end{figure}

\subsection{Inferring one-dimensional network dynamics}
{To comprehensively verify the effectiveness of the LLC, we test on six representative one-dimensional homogeneous network dynamics models, including Biochemical (Bio) \cite{voit2000computational}, Gene regulatory (Gene) \cite{mazur2009reconstructing}, Mutualistic Interaction (MI) \cite{gao2016universal}, Lotka-Volterra (LV) \cite{macarthur1970species}, Neural (Neur)\cite{wilson1972excitatory}, and Epidemic (Epi) \cite{pastor2015epidemic}, which have widespread applications across various fields, including biology, ecology, epidemiology, genomics, and neuroscience, exhibiting diverse characteristics.} 
{All network dynamics are simulated on an Erdős-Rényi (ER) network \cite{erdds1959random}}, where node degrees are drawn from a Poisson distribution with an average degree of $k=(N-1)p$ and $p$ is the probability of edge creation, so as to produce the continuous system states, i.e., $X(.)$. 
We apply the LLC to reconstruct governing equations from the observations $O=\{(X(t),A,M_x,M_a)\}_{t=0}^{T}$ sampled from $X(.)$.
Please see Appendix \ref{secA1} for the detailed description of the network dynamics models and specific settings.




We compare against the state of the arts for interpretable dynamics learning in complex networks, i.e., TPSINDy \cite{gao2022autonomous} and GNN+GP \cite{cranmer2020discovering,liu2023we}.
Due to the need for manual setting of function terms for the TPSINDy, we have set up various libraries with prior assumptions to obtain more comprehensive testing.
For the GNN+GP, we use an empirically suitable graph neural network version for simulating network dynamics \cite{liu2023we}, and a high-performance genetic programming-based symbolic regression, i.e., PySR \cite{cranmer2020pysr}, to output equations subsequently.
Also, we feed the output of our LLC as the discovered function terms into the TPSINDy, named as LLC+TPSINDy, to overcome its limitation of requiring a pre-defined library and verify our ability to discover new knowledge, i.e., effective function terms.
Fig.~\ref{fig2}(a) shows the comparison of the accuracy on predictions ($R^2$ score) and discovered equations (Recall) for reconstructing dynamics from six scenarios.
We see that the choice of function terms highly dominates the TPSINDy's results.
The TPSINDy with a library that has the same function terms as ground truth (TPSINDy-$\mathcal{H}_{N}$\footnote{$\mathcal{H}_{N}$ is a library that precisely contains the function terms of the objective equation.}) outperforms those with the libraries of only having basic operations (TPSINDy-$\mathcal{H}_{B}$\footnote{$\mathcal{H}_{B}$ only contains basic operations, such as polynomials, trigonometry, fractions, and exponential functions.}), lacking ground truth terms (TPSINDy-$\mathcal{H}_{W}$\footnote{$\mathcal{H}_{W}$ lacks the function terms of the objective equation.}), and containing redundancy terms (TPSINDy-$\mathcal{H}_{R}$\footnote{$\mathcal{H}_{R}$ includes not only $\mathcal{H}_{N}$ but also $\mathcal{H}_{B}$.}).
Compared to methodologies on the TPSINDy, the GNN+GP can achieve high-precision state prediction and equation reconstruction in more scenarios, i.e., LV, Epi, and Bio.
In contrast, our LLC achieves optimal performance covering all network dynamics scenarios and significantly outperforms the comparative methods in all metrics. 
More comparison results for other metrics, including MRE (Mean Relative Error), MAE (Mean Absolute Error), L2 error, and Precision, as well as the discovered equations for each scenario can be found in Appendix \ref{secA2}.
Note that the LLC+TPSINDy shows comparable performance and significantly improves TPSINDy, demonstrating that our LLC has discovered effective function terms to enhance existing methods.
We also report the average time required for different symbolic regression methods to regress the expected equations in all scenarios, as shown in Fig.~\ref{fig2}(b).
Undoubtedly, genetic programming-based PySR requires substantial computational time, i.e., $\sim$10 minutes, to carefully assemble and search expression trees.
The introduction of prior function terms and the execution of simple linear regression have brought high efficiency to the TPSINDy, which takes only about 5 minutes, reducing time costs by 50\%.
It is reasonable that transformer-based pre-trained methods, including symformer \cite{symformer}, E2E-ref \cite{e2eref}, and NSRA \cite{biggio2021neural}, have relatively low time costs, as they only perform efficient forward propagation once and optional constant optimization.
Note that the NSRA has not only high efficiency for equation regression, i.e., $\sim$4.5 minutes, but also has high accuracy, where the effectiveness of the discovery equations for each symbolic regression method is shown in Fig.~\ref{fig2}(c-e).
This is why we chose the NSRA as our symbolic regression module in the proposed pipeline.
{Despite the introduction of additional training computation for decoupling signals, i.e., $\sim$3 minutes, it achieves a good balance between efficiency and accuracy.}

{Moreover, we add Gaussian noise into node states to evaluate the robustness of our tool, where the amount of noise is measured by the signal-to-noise ratio (SNR) \cite{johnson2006signal}.}
As the SNR drops from 70 dB to 30 dB, ours always maintains an MSE close to 0, while more noises weaken performance but still significantly outperforms the TPSINDy (Fig.~\ref{figc1}(a)), which possibly is a benefit brought by neural network fitting in signal decomposition.
Also, we randomly delete edges on the network to simulate situations where the topology is difficult to fully obtain.
Fig.~\ref{figc1}(b) shows that the LLC can accommodate more missing edges.
The main reason is that it decouples the network dynamics, the missing edges do not affect the learning for one's self dynamics, and only affect the amount of data on the learning for interaction dynamics.
The optional constant optimization part in symbolic regression partially compensates for this.

\subsection{Inferring multi-dimensional and heterogeneous network dynamics}

\begin{figure}[t]
    \centering    
    \includegraphics[width=\textwidth]{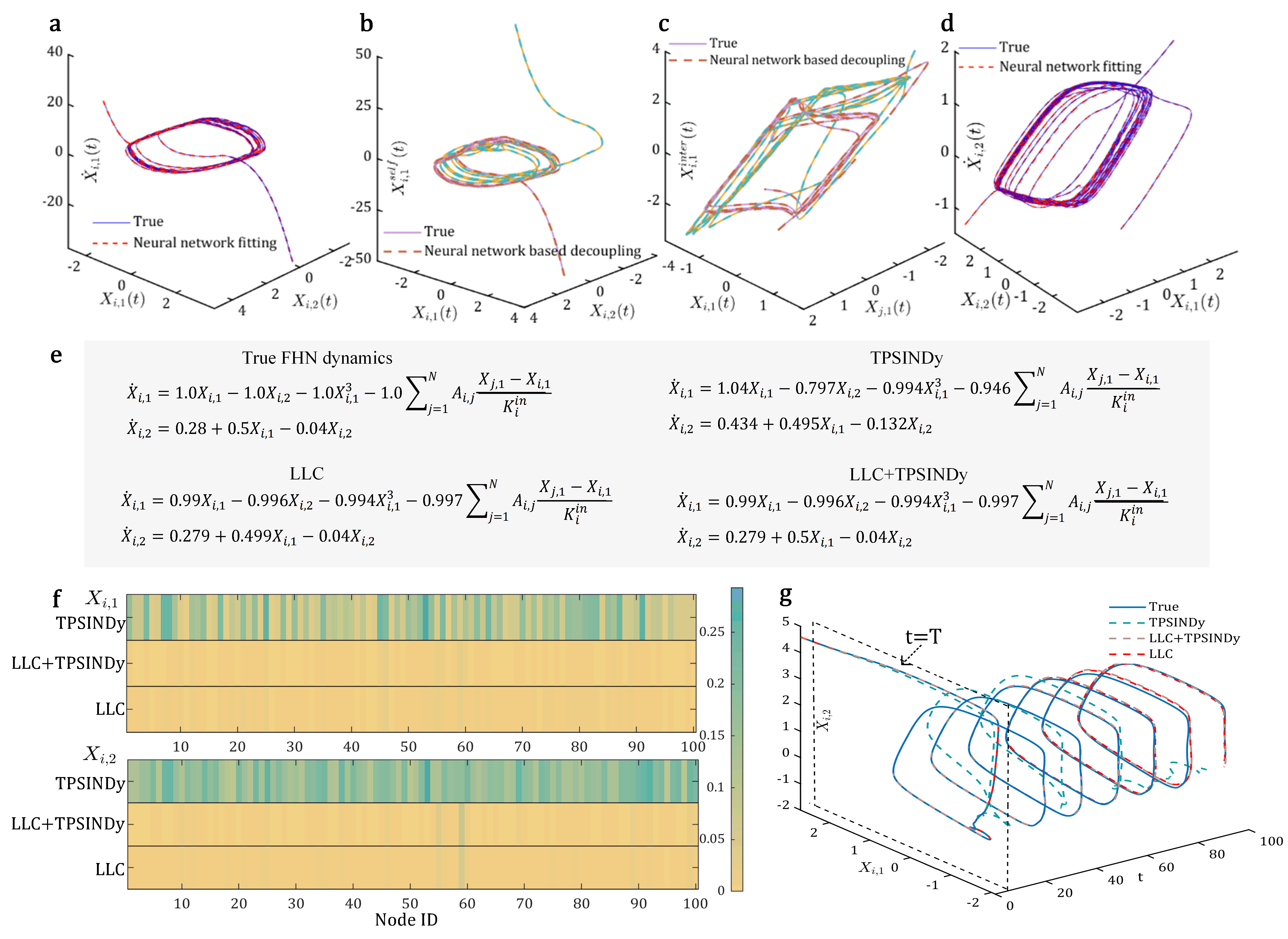}    
    \caption{
    {Results of inferring the FitzHugh-Nagum (FHN) dynamics. 
    \textbf{a.} The fitting results of the first dimension ($\dot{X}_{i,1}(t)$) on a node by neural networks.
    \textbf{b.} The decoupling results of the self dynamics for the first dimension on a node ($\Hat{Q}_{\theta_1}^{(self)}$).
    \textbf{c.} The decoupling results of the interaction dynamics for the first dimension on a node ($\Hat{Q}_{\theta_2}^{(inter)}$).  
    \textbf{d.} The fitting results of the second dimension ($\dot{X}_{i,2}(t)$) on a node by neural networks.
    \textbf{e.} Comparison of governing equations inferred by various methods.
    \textbf{f.} Comparison of the normalized Euclidean distance (NED) between two trajectories generated separately from the inferred and true equations, where the horizontal axis represents the node index.
    \textbf{g.} Comparison of trajectories generated by the inferred and true equations on a Barabási-Albert network.}
    } 
    \label{fig3}
\end{figure}

We apply the LLC to two multi-dimensional systems including FitzHugh-Nagumo (FHN) dynamics \cite{fitzhugh1961impulses} and predator-prey (PP) system \cite{fetecau2011swarm}.
The FHN is a two-dimensional neuronal dynamics that brain activities governed by the FitzHugh-Nagumo model, capturing the firing behaviors of neurons with two components, i.e., membrane potential ($X_{i,1}(t)$) and recovery variable ($X_{i,2}(t)$), which has a quasi-periodic characteristic.
A Barabási-Albert (BA) network is used to simulate the interactions among brain functional areas \cite{rabinovich2006dynamical}.
We get the observations of the sampling nodes by setting the masks, i.e., $M_x$ and $M_a$.
As shown in Fig.~\ref{fig3}(a-d), our LLC can effectively decompose the self dynamics and interaction dynamics by training neural networks $\Hat{Q}_{\theta_1}^{(self)}$ and $\Hat{Q}_{\theta_2}^{(inter)}$.
Moreover, compared to the TPSINDy, the equations discovered by the LLC and LLC+TPSINDy are closer to the true multi-dimensional governing equations (Fig.~\ref{fig3}(e)), and can achieve smaller predictive errors across all nodes (Fig.~\ref{fig3}(f)).
Especially as the prediction time increases ({up to $100T$}), our LLC can still make accurate trajectories with periodicity, while the TPSINDy finds it difficult to capture appropriate system behavior (Fig.~\ref{fig3}(g)), 
demonstrating that discovering true dynamic equations from data is crucial for improving long-term predictions.
{Moreover, we test our LLC on a Barabási-Albert network and two empirical networks, including the C. elegans \cite{yan2017network} and Drosophila \cite{scheffer2020connectome}, to analyze whether different topologies have an impact on the results.}
Fig.~\ref{figfhn} shows that it can obtain satisfactory equations and predictive trajectories on both synthetic and empirical topologies.
More settings can be found in Appendix \ref{secC}.

\begin{figure}[t]
    \centering   
    \includegraphics[width=\textwidth]{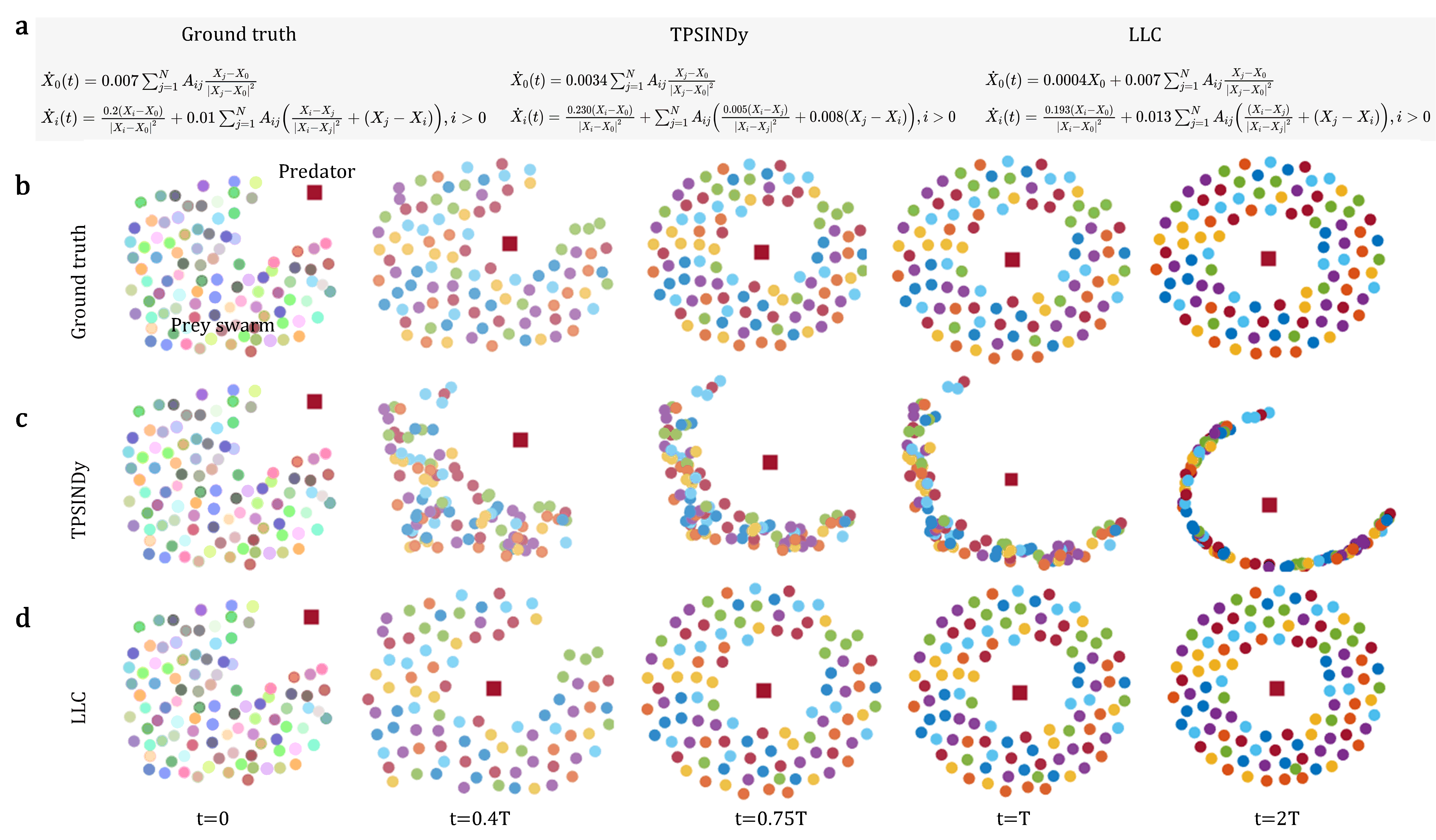}    
    \caption{
    {
    Results of inferring the predator-prey (PP) system.
    \textbf{a.} Comparison of governing equations inferred by various methods.
    \textbf{b.} The ground truth positions of a predator (square) and prey swarm (dots) over time.
    \textbf{c.} The predictive positions generated by the governing equation inferred by the TPSINDy.
    \textbf{d.} The predictive positions generated by the governing equation inferred by the LCC. 
    }
    }
    \label{fig4}
\end{figure}

The predator-prey (PP) is a heterogeneous system \cite{chen2014minimal}, where node state is the individual's position, but nodes are divided into two roles, i.e., {a predator ($i=0$) and many preys ($i>0$)}, resulting in three types of pair-wise relationships, i.e., predator-prey, prey-predator, and prey-prey.
Here, we consider a fully connected topology structure.
By setting masks $M_x$ and $M_a$ to form inputs for nodes with different types, dynamics equations are generated for the predator and prey respectively.
Note that due to the various types of nodes and edges, we set up multiple neural networks based on heterogeneous physical priors to decompose the mixed network dynamics signals, i.e., $\Hat{Q}_0^{(self)}$ and $\Hat{Q}_{0,j}^{(inter)}$ for predator, and $\Hat{Q}_i^{(self)}$, $\Hat{Q}_{i,j}^{(inter)}$, as well as $\Hat{Q}_{i,0}^{(inter)}$ for prey.
When testing, the prey are randomly dispersed while the predator enters from the top right corner as the initial system state.
Over time, the predator opens up a path within the prey swarm, ultimately forming a confused predator ring equilibrium state (Fig.~\ref{fig4}(b)).
The predator catches up with the swarm but is trapped at the center of the prey swarm and the swarm forms a stable concentric annulus around it, where the repulsion exerted by the predator cancels out due to the symmetry.
As shown in Fig.~\ref{fig4}(a), our LLC can discover a more accurate governing equation than the TPSINDy.
Although the TPSINDy produces the equation that matches the form of the true equation, there is a significant discrepancy between its predictive motion trajectories and ground truth, forming an unclosed ring with a large radius (Fig.~\ref{fig4}(c)).
In contrast, the equation discovered by our LLC indeed captures the same steady-state behavior of the swarm (Fig.~\ref{fig4}(d)).

\subsection{Inferring dynamics of chaotic systems}

\begin{figure}[t]
    \centering    
    \includegraphics[width=\textwidth]{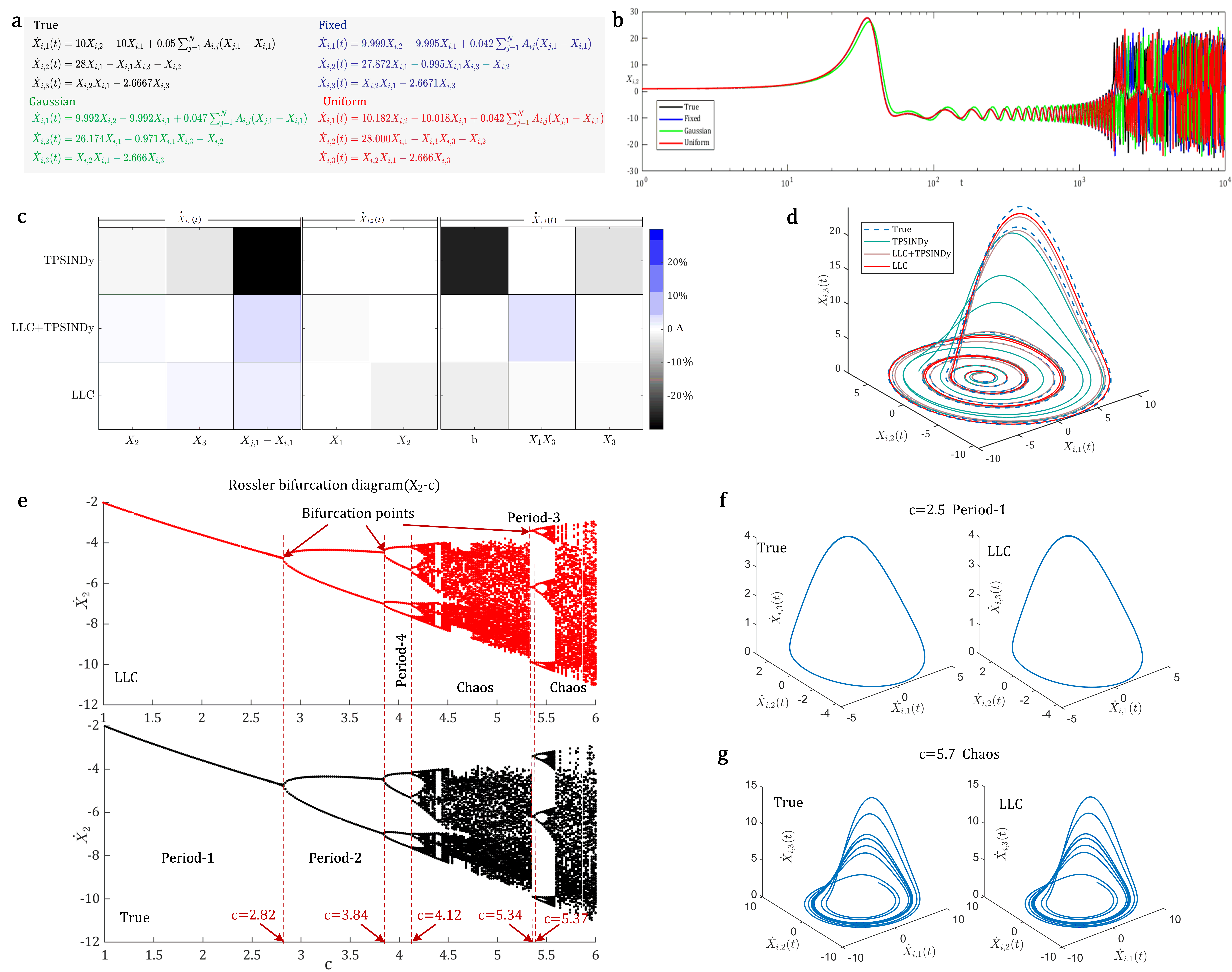}    
    \caption{
    Results of inferring the dynamics of chaotic systems.
    \textbf{a.} Comparison of governing equations inferred by our LCC under different initial conditions on the coupled Lorentz system.
    \textbf{b.} Comparison of predictive trajectories with the same initial values, produced by equations inferred by our LLC under different initial conditions.
    \textbf{c.} Coefficient errors between the equations inferred by each method on the Rossler system and the true equation.
    \textbf{d.} Comparison of trajectories on the same node, generated by the governing equations inferred by the TPSINDy, LCC, and LCC+TPSINDy on the Rossler system.
    \textbf{e.} Bifurcation diagram of the Rossler system via the Poincaré section method, with the horizontal axis representing the parameter $c$ ({ranging from 1 to 6}) and the vertical axis representing the system’s state on the second dimension ($X_{i,2}$) of a node. The discovered equation exhibits the same period-doubling and chaotic phenomenon as the true equation.
    \textbf{f.} Comparison of limit cycle at period-1, i.e., $c=2.5$.
    \textbf{g.} Comparison of chaos at $c=5.7$.
    }
    \label{fig5}
\end{figure}


We further discover the governing equations for three-dimensional chaotic systems on networks, including Lorenz \cite{MOGHTADAEI2012733} and Rössler dynamics \cite{xiao2007phase}.
To examine the impact of the initial sensitivity of chaotic systems on our LLC, we specify the following three initial value settings: fixing the initial states of all nodes to 0.1, sampling from a standard Gaussian distribution, and sampling from a uniform distribution U(0, 2).
Then, we employ a random network as the topological structure to produce the dynamics data of the Lorenz.
We see that the governing equations found from the dynamics data generated from different initial values are very similar (Fig.~\ref{fig5}(a)).
The predictive trajectories of the three discovered equations starting from the same initial values are close to the actual system behavior before 1000 time steps, and then the butterfly effect begins to gradually appear afterward (see Fig.~\ref{fig5}(b)), which means that our tool can accurately forecast a chaotic system for a long time, that is, 1000 times the training time steps.
By comparing the TPSINDy on the Rössler system, our LLC restores more accurate governing equations (Fig.~\ref{fig5}(c)) and has smaller predictive errors (Fig.~\ref{fig5}(d)).
We also analyze the state transition behavior of the Rössler system, and it is evident from the comparison of bifurcation diagrams generated by the inferred and true equations that period-doubling patterns and the bifurcation points are consistent, showing the transition process from period-1 to period-2, to period-4, to chaos, then to period-3, and finally to chaos again (see Fig.~\ref{fig5}(e)).
Fig.~\ref{fig5}(f-g) also shows the same limit cycle at period-1 and chaotic behavior.
This confirms the potential of our tool in analyzing the complex behavior and properties of systems with unknown dynamics.
More settings, experimental results, and analysis on chaotic systems can be found in Appendix \ref{secD}.

\subsection{Inferring dynamics of empirical systems}

\begin{figure}[t]
    \centering    
    \includegraphics[width=\textwidth]{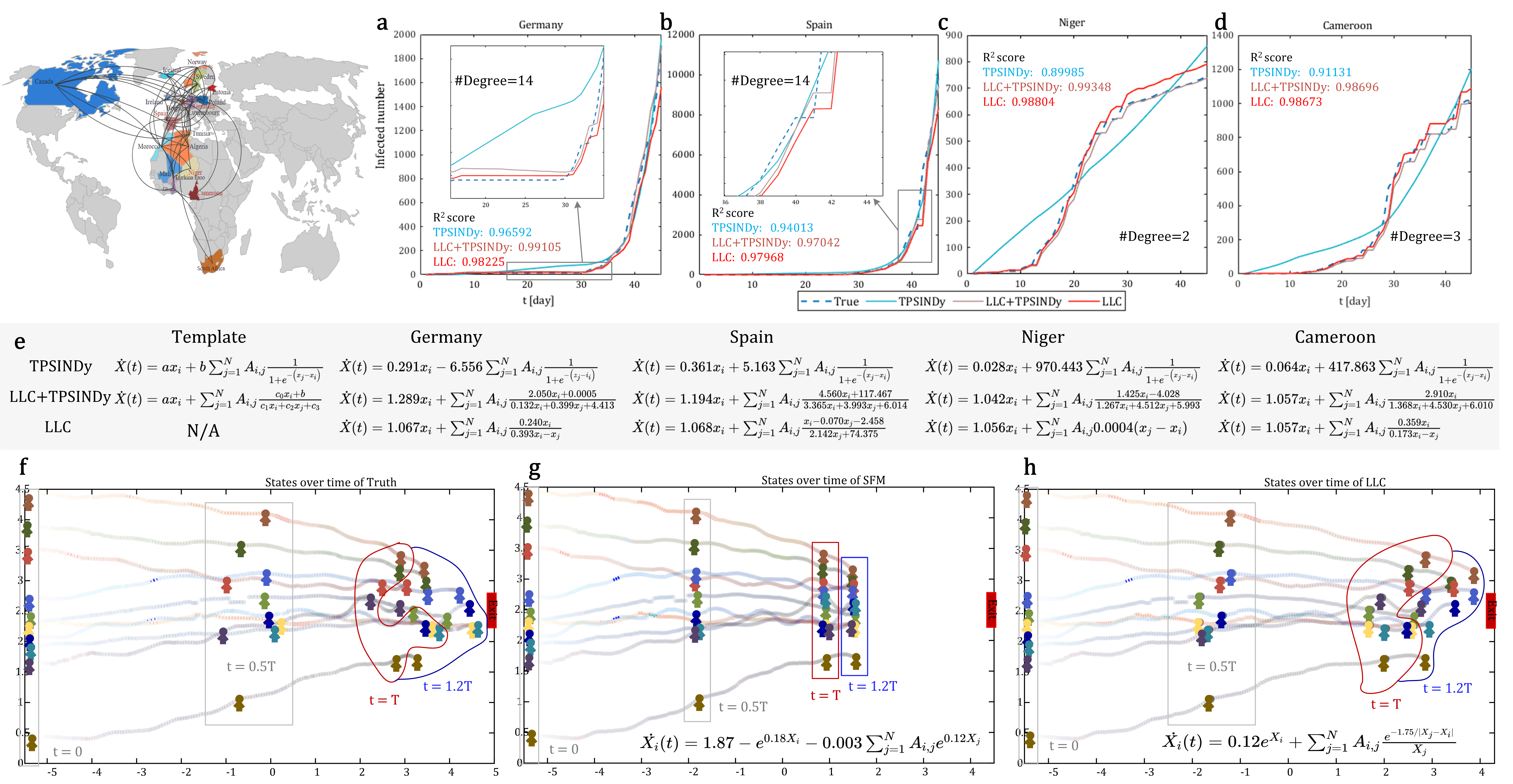}    
    \caption{
    Results of inferring the dynamics of empirical systems, including global COVID-19 transmission and pedestrian dynamics.
    \textbf{a-d.} Comparison of the number of cases over time in various countries or regions generated by TPSINDy, LLC, and LLC+TPSINDy.
    \textbf{e.} Comparison of governing equations inferred by various methods for four representative countries or regions.
    Note that the template in the LCC+TPSINDy is the induction of equations generated by our LLC for each node.
    \textbf{f.} The real pedestrian movement trajectories over time, where the data before $T$ is used for learning. 
    \textbf{g.} The inferred results on the pedestrian dataset using a mainstream social force model (SFM). 
    \textbf{h.} The inferred results on the pedestrian dataset using our LLC.}
    \label{fig6}
\end{figure}

{As impressive network dynamics, we collect daily global spreading data on COVID-19 \cite{dong2020interactive}, and use the worldwide airline network retrieved from OpenFights \cite{openflights} as a directed and weighted empirical topology to build an empirical system of real-world global epidemic transmission.}
Only early data before government intervention, i.e., the first 50 days, is considered here to keep the spread features of the disease itself.
By comparing the number of cases in various countries or regions (Fig.~\ref{fig6}(a-d)), we found that 
the trend of the results predicted by the TPSINDy is relatively good in countries with many neighbors, while for countries with few neighbors, the results are overestimated in the early stages of disease transmission and underestimated in the later stages.
More results on other countries or regions can be found in Appendix \ref{secE}.
From the perspective of the discovered equations (Fig.~\ref{fig6}(e)), due to the bounded interaction dynamics term in the TPSINDy, where each neighbor gives a maximum variation of 1, it has been providing a linear increase of approximately $kb$, where $k$ is the number of neighbors and $b$ is the coefficient of the interaction term, leading to overestimation in the early stage, while in the later stage, the linear increase of neighbors is insufficient and the self dynamics term is small, resulting in underestimation.
Counterintuitively, as the node degree decreases, $a$ decreases while $b$ increases in the TPSINDy, indicating that the growth of cases is more influenced by neighbors than oneself when the number of neighbors decreases. 
When predicting the case number in Spain by the TPSINDy, $b$ appears to be negative, meaning neighbors are always impractical negative contributions to oneself.
By comparison, predictive curves and the equations generated by our LLC and LLC+TPSINDy are more reasonable and accurate, with $R^2$ score closer to 1.
The change patterns discovered by ours focus more on the influence of their self dynamics and have smaller interaction terms.
As the node degree decreases, $a$ increases, that is, the influence of neighbors decreases while the influence of oneself increases, which is intuitive.
Moreover, the interaction term we discovered is similar to that of the mutualistic dynamics in ecosystems \cite{gao2016universal}, and conforms to the law of population evolution, especially in the equations discovered by the LLC+TPSINDy.
The above results demonstrate that we may have discovered a more suitable transmission model for the global disease.

For the other real-world system, we collect the crowd movement trajectories on a group of people walking in the same direction through the corridor towards a fixed destination \cite{boltes2013collecting} and use the k-nearest neighbor method to construct the topological structure (k=3) as the underlying propagation topology.
By comparing with the mainstream social force models (SFM) \cite{helbing1995social} in pedestrian dynamics, the equation discovered by ours exhibits a phenomenon close to the ground truth, that is, the crowd gradually moves and gathers towards the destination, occurring dispersion, aggregation and queuing, while the learned SFM causes the crowd to move forward side by side, as shown in Fig.~\ref{fig6}(f-h).
Note that the core distinction between the discovered equation and the SFM lies in the introduction of pedestrian distance in the interaction term, which is in line with the famous territorial effect \cite{scheflen1976human} that pedestrians always hope to maintain a distance from others.
This also demonstrates the applicability of ours to real data with unknown pattern of changes.

\section{Discussion}\label{sec3}

We introduced a computational tool to efficiently and accurately discover the effective governing ordinary differential equations from observed data of network dynamics.
We proposed decoupling network dynamics signals through neural networks and physical priors, and parsing governing equations via a pre-trained symbolic regression, to alleviate the challenges of excessive free variables and high computational cost.
Through sufficient experiments on various network dynamics from biology, ecology, epidemiology, genomics, and neuroscience, and comparing against the latest methods, our LLC can reconstruct the most accurate governing equations with tolerable execution efficiency, even in the face of noisy data and incomplete topologies.
It also performs well in complex situations, i.e., multi-dimensional and heterogeneous systems, such as brain dynamics and predator-prey system, demonstrating its universality and practicality.
Interestingly, by analyzing the behavior of chaotic systems and recovering bifurcation diagrams, we demonstrate that it can use the time series from unknown dynamics to gain insight into its properties, such as the existence of phase transitions.
We also show how it can deal with real-world datasets, including global epidemic transmission and pedestrian movements, and have actually discovered more effective symbolic models.
In a way, we view our tool as the equivalent of a numerical Petri dish to gain insights about unidentified dynamics behind observations, advancing the development for breaking through the inexplicable change mechanisms of complex phenomena.

Although the developed tool provides a feasible solution for the analysis and interpretability of complex systems, how to acquire the underlying topology structures remains an open question.
Moreover, many complex systems in real life are not purely autonomous systems, where external disturbances may also change over time \cite{wuprometheus}, exceeding the application boundary of this work.
More complex application scenarios are also challenges, such as non-deterministic systems \cite{stochasticNC2024}, partial differential systems \cite{raissi2018hidden}, and high-order network interactions \cite{battiston2021physics}.
Another interesting combination with large language models (LLM) is to become an external tool for LLM-based artificial intelligence agents to assist in reasoning and decision-making, by enabling efficient discovery of patterns of changes from data.
Although there is still some room to go in achieving reality-centric artificial intelligence, we believe that developing practical tools, such as the one presented here, can prepare us for more complex and urgent phenomena such as outbreaks of new infectious diseases and abnormal climate change.

\section{Methods}\label{sec4}

\subsection{Details of the LLC}
In this section, we present the various modules in our LLC, including data acquisition, selection of intervals, differential technology, physical priors, signal decoupling, symbolic regression, and termination conditions.


\textbf{Data acquisition}.
Generating synthetic data is a crucial step in validating algorithm performance and simulating real-world network behavior.
For a specific dynamics scenario $f$, we determine its hyperparameters and then choose a reasonable topology structure $A$, such as the Erdős-Rényi (ER) network \cite{erdds1959random}, C.elegans \cite{yan2017network}, and Drosophila \cite{scheffer2020connectome}.
After the initial state $X(0)$ is given, we can solve the Initial Value Problem (IVP) via the Runge-Kutta method \cite{cartwright1992dynamics} to simulate the network behavior at any time $t$, i.e., $X(t)$.
Note that we solve the IVP over a fixed time interval $t \in [0,T]$ and set both relative error and absolute error thresholds to $10^{-12}$ to ensure high precision throughout calculations which is a critical requirement in scientific research where accuracy of results holds paramount importance. 
Please refer to Appendix \ref{secA1} for detailed experimental settings of each scenario.
Additionally, we collect real data for the empirical systems, i.e., global spreading data \cite{dong2020interactive} and the trajectories of pedestrians \cite{boltes2013collecting}. 
More details can be found in Appendix \ref{secE}.

\textbf{Selection of intervals}.
{
When inferring the governing equations of complex network dynamics, 
whether the data has a reasonable sampling range, i.e., $[T', T]$, and the sampling interval, i.e., $\delta_t$, have a significant impact on the inference results.
For example, the data comes from all nodes reaching a consistent steady state is almost useless information for inferring the governing equation.
If the system exhibits periodic changes, the data we choose needs to contain at least one full period to absorb its periodic behavior.
For chaotic systems, we should have a sufficiently long time to observe the system's long-term behavior and stability, due to the sensitivity to initial conditions. 
Therefore, how to choose the reasonable range and interval is a practical problem that needs to be faced and solved.
To simplify the optimization process, we set the range to start from 0, i.e., $T'=0$, and establish the relationship between $T$ and $\delta_t$ as:
$
\delta_t=\frac{T}{S},
$
where $S$ is a given sampling steps.
Here, we employ the simulated annealing algorithm \cite{kirkpatrick1983optimization} to perform univariate optimization on $T$.
The objective function is as follows:
\begin{equation}
    T^*={\arg\min}_{T>0}\sum_{i=0}^{N-1}{\sum_{t=0}^{T-\delta_t}\left(\dot{X}_i(t+\delta_t)-\dot{X}_i(t)\right)^{2}+\lambda \sum_{t=T-10\delta_t}^{T}\left(\dot{X}_i(t)-\dot{X}_i(T)\right)^{2}},
\end{equation}
where $n$ is the number of nodes, $\lambda$ is an adjustable coefficient.  
Note that the first term on the right-hand side of the equation is to control that no drastic changes occur and the second term is to control the system to reach a steady state.
The objective not only minimizes amplitude fluctuations between adjacent periods but also ensures that changes within the total period remain as stable as possible. 

\textbf{Differential technology}. 
As a core and commonly used preprocessing technique for identifying ordinary differential equations \cite{brunton2016discovering,gao2022autonomous}, numerical differentiation can effectively calculate approximate derivatives, i.e., $\dot{X}(.)$, from time series of state data, i.e., $X(.)$.
The derivatives of measurement variables provide the targets for decoupling network dynamics signals by learning neural networks, i.e.,
$
\dot{X}_i(t)=Q_i^{(self)}(X_i(t))+\sum_{j=1}^{n}{A_{i,j}Q_{i,j}^{(inter)}(X_i(t),X_j(t))}
$,
where the subscripts $i$ and $j$ represent the corresponding nodes, $A_{i,j}$ is the adjacency matrix, $Q_i^{(self)}$ and $Q_{i,j}^{(inter)}$ capture the evolution process of their own states and the neighbors' dynamical mechanism governing the pairwise interactions respectively.
However, the inadvertent calculation of these derivatives may directly suffer the quality of the learned neural networks.
Conventional finite-difference approximations will greatly amplify any noise present in the state data; that is, even a small amount of noise can lead to a noticeable degradation in the quality of the numerical derivatives \cite{Numerical_differentiation}.
Smoothed finite difference is an effective way to mitigate the effect of noise through smoothing the time series data before calculating the derivative \cite{de2020pysindy}, which is also a widely applicable technology that has been empirically validated and recommended for use in identifying ordinary differential equations \cite{brunton2016discovering,de2020pysindy,gao2022autonomous}.
Therefore, we use this method to carefully calculate approximate derivatives, avoiding noise as much as possible.
Specifically, we use the {Savitzky-Golay filter \cite{liu2016applications}} to smooth the original state time series data and then approximate the derivatives through a five point finite difference method as follows: 
\begin{equation}
\label{eq2}
    \dot{{X}}_{i}(t) \approx \frac{{X}_{i}(t-2\delta_t)-8 {X}_{i}(t-\delta_t)+8 {X}_{i}(t+\delta_t)-{X}_{i}(t+2 \delta_t)}{12 \delta_t},
\end{equation}
where $\delta_t$ is the time interval between data points.

\textbf{Physical priors}.
A widely recognized physical prior in network dynamics is that the change of network states can be influenced by its own states and the states of its neighbors \cite{barzel2013universality,gao2022autonomous,liu2023we,SCIS_ours},
which can be formalized as 
\begin{equation}
\label{eq3}
\dot{X}_i(t)=Q_i^{(self)}(X_i(t))+\sum_{j=1}^{N}{A_{i,j}Q_{i,j}^{(inter)}(X_i(t),X_j(t))},
\end{equation}
where the subscripts $i$ and $j$ represent the corresponding nodes, $A_{i,j}$ is the adjacency matrix, $Q_i^{(self)}$ and $Q_{i,j}^{(inter)}$ capture the evolution process of their own states and the neighbors' dynamical mechanism governing the pairwise interactions respectively.
With the appropriate choice of $Q_i^{(self)}$ and $Q_{i,j}^{(inter)}$, Eq.~\ref{eq3} allows for a broad description of complex system dynamics \cite{barzel2013universality,barzel2015constructing}, such as biochemical processes, birth and death processes, heat diffusion, neuronal dynamics \cite{rabinovich2006dynamical}, gene regulatory \cite{karlebach2008modelling}, as well as chaos \cite{sprott2008chaotic} and oscillation \cite{rodrigues2016kuramoto} phenomena.
For homogeneous dynamics, the self dynamics of all nodes remain consistent, and those of all pairwise interactions are also the same. 
That is to say, we can ignore the subscripts $i$ and $j$, and directly obtain unique $Q^{(self)}$ and $Q^{(inter)}$.
For heterogeneous dynamics, we can group $Q_i^{(self)}$ and $Q_{i,j}^{(inter)}$ based on heterogeneous topology with $K$ node types and $E$ edge types, as follows:
\begin{equation}
\label{eq_hete}
\dot{X}_i(t)=Q_k^{(self)}(X_i(t))+\sum_{e=1}^{E}{\sum_{j=1}^{N}{A_{i,j}^{e}Q_{e}^{(inter)}(X_i(t),X_j(t))}},
\end{equation}
where, $1\leq k\leq K$ and $A_{i,j}^{e}$ is an adjacency matrix that only contains edges with type $e$.
At this point, we need to learn $K$ self dynamics functions and $E$ interaction dynamics functions.
Note that for an extreme case where all nodes have different self dynamics, as long as the above equation satisfies $K=n$.

\textbf{Signal decoupling}.
To achieve signal decoupling using the physical priors, we parameterize ${Q}_i^{(self)}$ and ${Q}_{i,j}^{(inter)}$ using neural networks, i.e., $\Hat{Q}_i^{(self)}$ and $\Hat{Q}_{i,j}^{(inter)}$.
Specifically, we use a multi-layer perceptron to model the nonlinearity of the self dynamics. 
That is, $\Hat{Q}_{i}^{(self)}(X_i(t)):=\boldsymbol{\psi}^{f}(X_i(t))$ is a simple feed-forward neural network with $L$ hidden layers, where each hidden layer applies a linear transformation followed by a rational activation function.
In order to have the ability to characterize broad dynamics scenarios, we design a neural network that approximates the interaction dynamics as:
$
\Hat{Q}_{i,j}^{(inter)}(X_i(t), X_j(t)) := \boldsymbol{\psi}^{g_0}(X_i(t), X_j(t))
    +[\boldsymbol{\psi}^{g_{1}}(X_i(t)) \times \boldsymbol{\psi}^{g_{2}}(X_j(t))],
$
where $\boldsymbol{\psi}^{g_0}$, $\boldsymbol{\psi}^{g_1}$ and $\boldsymbol{\psi}^{g_2}$ denote non-shared multi-layer perceptrons.
The first term on the right-hand side represents a coupled nonlinear interaction term, e.g., $(X_j-X_i)\times X_j$, while the second term represents a decomposable one, e.g., $X_j/(1+X_i^2)$.
This way, regardless of whether the interaction item is decomposable or not, it can be automatically learned.
Then, the optimal neural approximation can be obtained by minimizing the loss $\mathcal{L}$ between $\dot{X}(t)$ and the predictive $\hat{\dot{X}}(t)$ calculated by $\boldsymbol{\psi}^{f}$, $\boldsymbol{\psi}^{g_0}$, $\boldsymbol{\psi}^{g_1}$, and $\boldsymbol{\psi}^{g_2}$:
\begin{equation}
\label{eq6}
\begin{aligned}
\mathcal{L} =\frac{1}{N \times d}\left[||\dot{X}(t)-\hat{\dot{X}}(t)||_{1}
+\lambda \frac{1}{N-1}\sum_{i=1}^N(||\dot{X}_{i}(t)-\hat{\dot{X}}_i(t)||_1+\mathbb{E}[\dot{X}_{i}(t)-\hat{\dot{X}}_i(t)])^2\right],
\end{aligned}
\end{equation}
where $N$ is the number of nodes, $d$ is the state dimension, and $||.||_{1}$ is $\ell_1$ norm.
The first part of the loss function represents the mean absolute error and the second part represents the variance of the absolute error. 
The parameter $\lambda$ is a coefficient used for balance.
The loss function combines the mean and the variance, aiming not only to minimize the average difference between the predictions and the true values but also to control the fluctuation of the error, which can make the model more stable during the training process, avoiding overfitting or underfitting. 
Note that if the network dynamics are homogeneous, meaning that the governing equations for all nodes are consistent, then only two neural networks need to be learned, i.e., $\Hat{Q}_{\theta_1}^{(self)}:=\Hat{Q}_{i}^{(self)}$ and $\Hat{Q}_{\theta_2}^{(inter)}:=\Hat{Q}_{i,j}^{(inter)}$.
To accommodate the heterogeneous dynamics, i.e., Eq.~\ref{eq_hete}, 
we need to learn multiple neural networks, i.e., $\{\Hat{Q}_k^{(self)}\}_{k=1}^{K}$ and $\{\Hat{Q}_e^{(inter)}\}_{e=1}^{E}$.
These neural networks have the same architecture but do not share parameters.
Specific architectures and training setup can be found in Appendix \ref{sex_NN}.

\textbf{Symbolic regression}.
Symbolic regression is a supervised learning method that attempts to discover hidden mathematical expression $g: \mathbb{R}^{m}\to \mathbb{R}$ from a series of input-output points, i.e., $\{(x_i, y_i),...\}$,
where $x_i\in\mathbb{R}^{m}$ is input feature variables and $y_i\in\mathbb{R}$ is target variable. 
Transformer-based symbolic regression converts this task into a mapping process from data sequences to expression sequences.
A data sequence comprises many input-output pairs, while an expression sequence is the prefix expression \cite{biggio2021neural}.
A mathematical expression can be viewed as a tree, where intermediate nodes represent operators, while leaf nodes represent variables or constants \cite{LampleC20}.
Operators can be unary, e.g., $\sin$, $\cos$, $\tan$,  $exp$, or be binary, e.g., $+$, $-$, $\times$, $\setminus$.
Then, an expression tree can be converted into a corresponding expression sequence.
For example, $cos(2\times x_1)+3\times x_2$ is formatted as a prefix expression, i.e., $[+, cos, \times, 2, x_1, \times, 3, x_2]$. 
By tokenizing the expression sequences and constructing massive amounts of data and expressions, such as millions of samples, we can sufficiently pre-train a transformer in an end-to-end manner, integrating massive equation knowledge.
Only one forward propagation is required when inferring a new equation, significantly reducing the inference time.
Due to the unbounded nature of constants in expressions, tokenizing and accurately reconstructing constants is challenging,
Therefore, an optional adjustment constant stage can be added after the expressions generated by the pre-trained model to improve the accuracy of equation recovery and predictions.
Generally, Broyden–Fletcher–Goldfarb–Shanno (BFGS) \cite{fletcher2000practical} or stochastic gradient optimization \cite{bottou2010large} can be used to tune the constants in expressions to ensure inference accuracy.
Now, based on the inputs and outputs generated by trained neural networks, i.e., $\Hat{Q}_{\theta_1}^{(self)}$ and $\Hat{Q}_{\theta_2}^{(inter)}$, we can use the pre-trained symbolic regression model to parse their corresponding symbolic equations efficiently.

During pre-training, it is usually assumed that the data points are independent of time and sampled independently and identically.
However, in our network dynamics scenario, data points are time-dependent.
Consequently, directly feeding collected state trajectories into pre-trained symbolic transformers may not yield satisfactory regression results, as it has not been trained on such correlated data points.
Therefore, a suitable sampling strategy is needed to make the sampling data points as independent as possible.
One direct strategy is to sample with a uniform time step, but if the state trajectory contains rapidly changing segments, it may fail to accurately capture these behaviors, potentially resulting in uneven sample distribution.
{Hence, to enhance the effectiveness of the pre-trained symbolic transformer in identifying complex network dynamics, we employ the K-Means sampling method \cite{aggarwal2009adaptive} to obtain more representative data points.}
It can better adapt to changes in state trajectories, thereby improving sample diversity and distribution uniformity.

\textbf{Termination conditions}.
We can see from Fig.~\ref{fig1} that both signal decoupling and the overall discovery process require termination conditions. 
One for the signal decoupling part is to train the satisfactory neural networks to accurately decouple the self dynamics and interaction dynamics. 
Usually, we use whether the error of the validation set is less than a certain threshold to judge. 
For the termination conditions of the entire process, we determine whether the difference between the predictive results generated by the identified equation and the actual data is less than a certain threshold, and then judge whether a satisfactory dynamics equation is found.


\subsection{Performance measures}
The performance measures for evaluating the methods in this work are as follows:

\textbf{R$^2$ score}. $R^2$ score, also known as the coefficient of determination, 
is used to measure the predictive ability of a method, usually ranging from $-\infty$ to 1.
The closer the R$^2$ score is to 1, the better the prediction performance.
The score can be calculated as:
\begin{equation*}
    R^2=1-\frac{ {\textstyle \sum_{i=1}^{N}(X_i(t)-\hat{X}_i(t) )^2} }{\textstyle \sum_{i=1}^{N}(X_i(t)-\bar{X}(t) )^2},
\end{equation*}
where $X_i(t)$ and $\hat{X}_i(t)$ are the true and predictive states of node $i$ at time $t$, respectively. 
$\bar{X}(t)$ is the average of $X_i(t)$ over all node and $n$ is the number of nodes.

\textbf{NED}. We adopt the Normalized Euclidean distance (NED) to finely evaluate the difference between the predictive states produced by the inferred equation and the true states for each node $i$ as
\begin{equation*}
    \text{NED}(X_i,\hat{X}_i)=\frac{1}{D_{\text{max}}^{0:T}}{\textstyle \sum_{t=0}^{T}\sqrt{(X_i(t)-\hat{X}_i(t))^2+(\dot{X}_i(t)-\hat{\dot{X} }_i(t)  )^2} },
\end{equation*}
where, $D_{\text{max}}^{0:T}$ is the longest Euclidean distance in the pairs composed of the true states.
$X_i(t)$ and $\hat{X}_i(t)$ are the true and predictive states of node $i$ at time $t$, respectively.
$\dot{X}_i(t)$ and $\hat{\dot{X}}_i(t)$ are their respective derivatives.

\textbf{Recall}. From the perspective of the symbolic form of equations, we use recall to evaluate the accuracy of the discovered equations, i.e., whether a certain function term is present or not.
Let $\xi_{true}$ and $\xi_{pre}$ denote the true and inferred coefficients under a candidate library of function terms.
For example, a library is $[1, x, x^2, x^3, x^4]$,
If a true equation is $y=1+x$, then $\xi_{true}=[1,1,0,0,0]$.
And if an inferred equation is $y=1+x^2$, then $\xi_{pre}=[1,0,1,0,0]$.
Recall measures the percentage of coefficients successfully identified among the true coefficients and is defined as: 
\begin{equation*}
    Recall=\frac{| \xi_{true}\odot \xi_{pre}   |_0 }{|\xi_{true}|_0} ,
\end{equation*}
where $\odot$ represents the element-by-element product of two vectors, $|\cdot|_0$ denotes the number of non-zero elements in the vector.

{More performance measures, including MRE (Mean Relative Error), MAE (Mean Absolute Error), L2 error, and Precision, can be found in Appendix \ref{secA2}.}

\bibliography{sn-article}


\newpage

\begin{appendices}

\section{Network Dynamics and Topologies}\label{secA1}
\subsection{Network Dynamics}

We study six representative one-dimensional homogeneous network dynamics spanning the fields of biology, ecology, epidemics, and neuroscience.

\begin{itemize}
\item Biochemical Dynamics (Bio): Biochemical processes within living cells are mediated by protein-protein interactions in which proteins bind to form protein complexes \cite{voit2000computational}. Its dynamics of biochemical reactions can be formulated as: $\dot{X}_{i}(t)=F_{i}-B_{i}X_{i}(t)+\sum_{j=1}^N A_{i,j} X_{i}(t)X_{j}(t)$, where $X_{i}(t)$ is the concentration of protein ${i}$ at time $t$, $A_{i,j}$ is the effective rate constant of interaction between $i$ and $j$, and $F_{i}$ and $B_{i}$ represent the average influx rate and degradation rate of proteins, respectively. 
We set $F_{i} = 1$ and $B_{i} = -1$.
\item Gene Regulatory Dynamics (Gene): The dynamics of gene regulatory networks can be described by the Michaelis-Menten equation \cite{mazur2009reconstructing,karlebach2008modelling}, given as 
$\dot{X}_{i}(t)=-B_{i}X_{i}(t)^{f}+\sum_{j=1}^N A_{i,j} \frac{X_{j}(t)^{h}}{X_{j}(t)^{h}+1}$. The node state $X_{i}(t)$ is the expression level of gene $i$. 
Parameter $B_{i}$ controls the decay rate. 
When $f = 1$, the first term on the right-hand side describes degradation.
When $f = 2$, it describes dimerization corresponding to two or two of the same molecules together to form dimers.
The second term captures genetic activation, where $h\ge0$ represents the Hill coefficient, quantifying the saturation rate affected by neighboring nodes. We set $B_i = 2$, $f = 1$, and $h = 2$ here.
\item Mutualistic Interaction Dynamics (MI): The mutually beneficial interaction dynamics between species in ecology, governed by
$\dot{X}_{i}(t)=b_i+X_{i}(t)(1-\frac{X_{i}(t)}{k_i})(\frac{X_{i}(t)}{c_i}-1)+\sum_{j=1}^{N} A_{i,j} \frac{X_{i}(t) X_{j}(t)}{d_i+e_iX_{i}(t)+h_iX_{j}(t)}.$
The abundance $X_{i}(t)$ of a captured species in a mutualistic differential equation system \cite{gao2016universal} consists of an afferent migration term $b_i$, a logical increase in population capacity  $k_i$, an Allee effect with a cold starting threshold $c_i$, and a mutualistic interaction term with the interaction network $A$. The parameters are set as $b_i = 1$, $k_i = 5$, $c_i = 1$, $d_i = 5$, $e_i = 0.9$, and $h_i = 0.1$.

\item Lotka-Volterra Model (LV): The Lotka-Volterra model (LV) \cite{macarthur1970species} describes the population dynamics of species in competition: $\dot{X}_{i}(t)=X_{i}(t)(\alpha_{i}-\theta_{i}X_{i}(t))- \sum_{j=1}^{N} A_{i,j}X_{i}(t)X_{j}(t)$. Similar to mutualistic dynamics, the node state $\mathbf{x}_{i}(t)$ represents the population size of species $i$, the growth parameters of species $i$, $\alpha_{i},\theta_{i} > 0 $. In experiments, we sample $\alpha_{i},\theta_{i}$ from a uniform distribution within the range $[0.5, 1.5]$. Specifically, $\alpha_{i}$ and $\theta_{i}$ are set to 0.5 and 1 respectively.

\item Neural Dynamics (Neur): The firing rate of a neuron can be described by the Wilson-Cowan \cite{wilson1972excitatory} model as $\dot{X}_{i}(t)=- X_{i}(t)+\sum_{j}^{N}A_{i,j}(1+exp(-\tau(X_{j}(t)-\mu)))^{-1}$. In this model, $X_{i}(t)$ is the activity level of neuron $i$, while the parameters $\tau$ and $\mu$ determine the slope and threshold of the neural activation function, respectively. We set $\tau = -1$ and $\mu = 1$.

\item Epidemic Dynamics (Epi): Epidemic dynamics can be used to describe the outbreak of infectious diseases \cite{pastor2015epidemic} as: $\dot{X}_{i}(t)=-\delta_{i}X_{i}(t)+\sum_{j=1}^{N}A_{i,j}(1-X_{i}(t))X_{j}(t)$. In this model, each node can represent an individual, where the node state $X_{i}\in [0, 1]$ corresponds to the infection probability of node $i$. The parameter $\delta_{i}$ represents the rate at which individuals recover from infection, which is set to be 1.0.

\end{itemize}

\subsection{Network topologies}

Network topology affects the evolution of node states. 
In this work, we study four network topologies, including the Erdős-Rényi (ER) \cite{erdds1959random}, Barabási-Albert (BA) \cite{barabasi1999emergence}, and two empirical networks, i.e., the C.elegans \cite{yan2017network} and Drosophila \cite{scheffer2020connectome}.
Below we describe the unique generation process of each topology. 
 \begin{itemize}
     \item Synthetic networks: 
     (1) Erdős-Rényi (ER) network \cite{erdds1959random}, also known as a random network, where node degree is drawn from a Poisson distribution with mean degree $k=(n-1)p$. $p$ is the probability of edge creation.
     (2) Barabási-Albert (BA) network \cite{barabasi1999emergence}, characterized by a power-law degree distribution. We use a priority connection to construct the network topology, starting with a set of nodes and iteratively introducing new nodes. Each new node connects to existing nodes, the probability of connecting to a specific existing node is proportional to the degree of the node. 
     \item Empirical networks: (3) the neural connectome of the C.elegans \cite{yan2017network}, the nematode worm Caenorhabditis elegans, defined as the 279 neurons between 2990 synaptic connections on the cellular network. 
     (4) Drosophila \cite{scheffer2020connectome}, the cellular connectome of the mushroom body region of the fly, which includes neurons that project their axons into bundles similar to paired mushrooms, can be accessed through \url{https://neuprint-examples.janelia.org/}.
 \end{itemize}

The ER network is mainly used for data generation of comprehensive comparative experiments on one-dimensional homogeneous network dynamics, while the BA network and two empirical networks are mainly used for data generation when inferring multi-dimensional, heterogeneous, and chaotic network dynamics.

\subsection{Initial conditions, sampling intervals, and end time of inferring and predicting}
\setcounter{table}{0}

To ensure the reproducibility of our findings, we provide a comprehensive list of parameter settings on the initial conditions, sampling intervals, and end time of inferring and predicting, where let $T$ and $T_{end}$ denote the end time of inference and end time of prediction, respectively.
If not explicitly stated, the number of nodes ($N$) is set to 100.
The simulations encompass dynamics from time $t=0$ to $T$ or $T_{end}$ with a step-size $\delta t$.
Table \ref{tab:initial_conditions} shows the initial conditions, sampling intervals, and end time of inferring and predicting used to simulate network dynamics data.
\begin{table}[h]
\centering
\caption{Initial conditions, sampling intervals, and end time of inferring ($T$) and predicting ($T_{end}$) used to simulate network dynamics data.}
\label{tab:initial_conditions}
\begin{tabular}{crrrrr}
\toprule
Dynamics    & Initial Condition ($t=0$) & $\delta t$  & $T$ & $T_{end}$ \\
\midrule            
Bio & Uniform distribution(0,2)            & 0.0001  & 0.1  & 0.5 \\ 
Gene        & Uniform distribution(0,2)             & 0.01    & 5    & 10  \\
MI          & Uniform distribution(0,2)             & 0.001   & 1    & 5   \\
LV          & Uniform distribution(0,5)             & 0.0001  & 0.1  & 0.5 \\
Neur      & Uniform distribution(0,2)             & 0.01    & 5    & 10  \\
Epi         & Uniform distribution(0,1)            & 0.001   & 1    & 5 \\ \bottomrule

\end{tabular}
\end{table}


 \section{ Details of Neural Architectures in Signal Decoupling}\label{sex_NN}
\renewcommand{\thetable}{B\arabic{table}}
The detailed architectures of the neural networks $\boldsymbol{\psi}^{f}$, $\boldsymbol{\psi}^{g_0}$, $\boldsymbol{\psi}^{g_1}$ and $\boldsymbol{\psi}^{g_2}$ in $\hat{Q}^{(self)}$ and $\hat{Q}^{(inter)}$ are shown in Table \ref{tab:architectures}.
\begin{table}[h]
    \caption{The detailed architectures of the neural networks $\boldsymbol{\psi}^{f}$, $\boldsymbol{\psi}^{g_0}$, $\boldsymbol{\psi}^{g_1}$ and $\boldsymbol{\psi}^{g_2}$ in $\hat{Q}^{(self)}$ and $\hat{Q}^{(inter)}$, where $d$ is the state dimension of the system.}
    \centering
    \begin{tabular}{|c|c|c|c|c|}
        \hline Neural networks & $\boldsymbol{\psi}^{f}$ & $ \boldsymbol{\psi}^{g_0}$ &$\boldsymbol{\psi}^{g_{1}}$ & $\boldsymbol{\psi}^{g_{2}}$ \\
 \hline\multirow{6}{*}{Layers } & Linear($d$,50) & Linear($2d$, 50) & Linear($d$,50) & Linear($d$, 50)\\
 & Rational & Rational & Rational & Rational \\
 & Linear(50, 50) & Linear(50, 50) & Linear(50, 50) & Linear(50, 50) \\
 & Rational & Rational & Rational & Rational  \\
 & & Linear(50, 50) & Linear(50, 50) & Linear(50, 50)\\
 & & Rational & Rational & Rational\\
\hline Readout layer & Linear(50, $d$) & Linear(50, $d$) & Linear(50, $d$) & Linear(50, $d$) \\
\hline
\end{tabular}
\label{tab:architectures}
\end{table}
They were trained using a random division of the timestamps into training and validation sets, {with ratios of 0.8 and 0.2}, respectively, and trained for 1,000 epochs using the AdamW optimizer. 
When the error on the validation set is less than a certain threshold, the optimization process stops or stops early when multi-step validation error remains unchanged.
The learning rate was searched in the range of [1e-3, 1e-2], the weight decay value was set to 0.001, and the hidden dimension was set to 50.
Note that, we employ a smooth, trainable activation function derived from the rational function in Table \ref{tab:architectures}.
There are practical and theoretical reasons for choosing rational functions.
{First, rational functions are computationally efficient, mainly because the polynomials connected by the multivariate operators are well suited for parallel computation.}
Second, rational functions can approximate a wider range of functions more efficiently and accurately than polynomials \cite{walsh1935interpolation}, e.g., functions with singularities or sharp changes. 
The specific rational activation function can be formulated as:
\begin{equation*}
    R(X_i(t))=\frac{P(X_i(t))}{Q(X_i(t))} =\frac{ {\textstyle \sum_{i=0}^{r_P}a_i} X_i(t)}{1+|{\textstyle \sum_{j=0}^{r_Q}b_j} X_j(t)|},
\end{equation*}
where $P$ and $Q$ represent two polynomials, the numerator and denominator respectively.
$r_P$ and $r_Q$ are set to be 3 and 2, respectively.
The specific values of a and b are set as: $a_0=1.1915, a_1=1.5957, a_2=0.5000, a_3=0.0218$; $b_0=2.3830, b_1=0.0000, b_2=1.0000$.

\section{More Results of Inferring One-Dimensional Network Dynamics}\label{secA2}
\renewcommand{\thetable}{C\arabic{table}}
\subsection{Comparison of discovered equations}

Table \ref{Tb2} shows the discovered equations for each network dynamics scenario,  demonstrating the ability of various methods to learn interpretable network dynamics.
TPSINDy-$\mathcal{H}_{B}$ with only the basic operations and 
TPSINDy-$\mathcal{H}_{W}$ with lacking the function terms of the ground truth
struggle to infer the correct form of the equations for all dynamics systems. 
Although TPSINDy-$\mathcal{H}_{N}$ with a library that has the same function terms as ground truth and TPSINDy-$\mathcal{H}_{R}$ with sufficient basis functions can partially restore the terms of the equation, they still fails to accurately infer the coefficients.
That is to say, improper basis library settings due to a lack of prior knowledge make it even more difficult for the TPSINDy to identify the correct dynamics equations.

On the contrary, the proposed LLC can directly infer the dynamics equations accurately from the node states. 
Additionally, it can further enhance the TPSINDy. 
When terms obtained through the LLC are used as the basis function terms for the TPSINDy, the sparse regression method can accurately identify the dynamics equations, which can be supported by comparing the discovered equations of TPSINDy-$\mathcal{H}_{N}$ and LLC+TPSINDy.

\begin{landscape} 
\renewcommand{\arraystretch}{1.0}
\begin{threeparttable}
\begin{longtable}{p{2cm}p{3.1cm}p{3cm}p{3.8cm}p{3.2cm}p{2.8cm}p{3.2cm}}        
\caption{Comparison of inferred equations of network dynamics.}\label{Tb2}\\
    \hline
    \multicolumn{1}{c}{\textbf{Methods}}  & \multicolumn{1}{c}{\textbf{Bio}} & \multicolumn{1}{c}{\textbf{Gene}} & \multicolumn{1}{c}{\textbf{MI}} & \multicolumn{1}{c}{\textbf{LV}} & \multicolumn{1}{c}{\textbf{Neur}} & \multicolumn{1}{c}{\textbf{Epi}} \\
    \hline
    True &
    $\begin{array}{l}
        1-X_{i}+\sum_{j=1}^{N} \\A_{i j}X_{i}X_{j}
     \end{array}$ & 
    $\begin{array}{l}
        -2.000 X_{i}+ \\\sum_{j=1}^{N} A_{i j} \frac{X_{j}^{2}}{1+X_{j}^{2}}
     \end{array}$ & 
    $\begin{array}{l}
        1+X_i(1-\frac{X_i}{5})(X_i-1) \\\sum_{j=1}^{N} A_{i j} \frac{X_{i} X_{j}}{5+0.9 X_{i}+0.1 X_{j}}
    \end{array}$ & 
    $\begin{array}{l}
        0.5X_{i}-X_{i}^{2}-\sum_{j=1}^{N} \\A_{i j} X_{j} X_{i}
    \end{array} $ & 
   $\begin{array}{l}
        -X_{i}+\sum_{j=1}^{N} A_{ij} \\\frac{1}{1+e^{-(X_j-1)}}
   \end{array} $ &   
   $\begin{array}{l}
        -X_{i}+\sum_{j=1}^{N} A_{i j}\\ X_{j}\left(1-X_{i}\right)
    \end{array} $ 
    \\ \hline
    TPSINDy$-\mathcal{H}_{B}$ &  
    $\begin{array}{l}
        0.004e^{X_i}+ \sum_{j=1}^{N} \\A_{i j} 0.494X_{j} X_{i}
     \end{array}$ & 
    $\begin{array}{l}
        -1.163 X_{i}+ \sum_{j=1}^{N} \\A_{i j}\left(2.008+0.026 X_{j}\right)
     \end{array}$ & 
    $\begin{array}{l}
        -6.407+ \\\sum_{j=1}^{N} A_{i j} \frac{77.520}{X_{i} X_{j}}
    \end{array}$ &
   $\begin{array}{l}
        -0.181 X_{i}^{2}-\\\sum_{j=1}^{N} A_{i j} 1.135 X_{j} X_{i}
    \end{array}$ & 
   $\begin{array}{l}
        -0.123X_{i}^{2}+8.352+\\\sum_{j=1}^{N} A_{i j} 1.128 \sin X_j
    \end{array}$ & 
   $\begin{array}{l}
        -0.769 X_{i}^{3}+ \sum_{j=1}^{N} A_{i j}\\1.330\left(X_{j}-X_{i}\right)
    \end{array}$ 
   \\ \hline
    TPSINDy$-\mathcal{H}_{N}$ & 
    $\begin{array}{l}
        0.078  X_{i}^{3}+ \\\sum_{j=1}^{N} A_{i j}0.923X_i X_j
    \end{array}$ & 
   $\begin{array}{l}
        2.071-0.853 X_{i}+ \\\sum_{j=1}^{N} A_{i j} 0.050  \frac{X_{j}^{2}}{1+X_{j}^{2}}
    \end{array}$ & 
   $\begin{array}{l}
        1.047 X_{i}-0.015 X_{i}^{3}+\\ {\sum_{j=1}^{N} A_{i j}}\\\left(0.01 X_{j}-0.281\left(X_{j}-X_{i}\right)\right)
    \end{array}$ & 
    $\begin{array}{l}
        -0.894-0.084 X_{i}^{2}\\-\sum_{j=1}^{N} A_{i j} 1.42 X_{j}X_{i}
    \end{array}$ & 
    $\begin{array}{l}
        -4.278 X_{i}+28.343+\\\sum_{j=1}^{N} A_{ij} \frac{-0.215}{1+e^{-(X_j-1)}}
     \end{array}$ & 
   $\begin{array}{l}
        4.126+ \sum_{j=1}^{N} A_{ij}\\ \left(-1.286 X_{j}-0.147X_{j} X_{i}\right)
    \end{array}$ 
    \\ \hline
    TPSINDy$-\mathcal{H}_{W}$ &  
    $\begin{array}{l}
        {2.125} \cos X_i+ \frac{311.387}{e^{-(X_{i}-5)}}\\+ \sum_{j=1}^{N} A_{i j}(\frac{3.151 X_{i} X_{j}}{X_{j}+1} \\+\frac{0.590(X_{j}-X_{i})}{K_{\text {in }}})
    \end{array}$ & 
   $\begin{array}{l}
        \frac{1.392 X_{i}}{K_{\text {in }} }+ \sum_{j=1}^{N} \\A_{i j} \frac{1.204\left(X_{j}-X_{i}\right)}{K_{\text {in }}}  
    \end{array}$ & 
   $\begin{array}{l}
        \frac{-81.872}{e^{\left(-\left(X_{i}-10\right)\right)}}+ {\footnotesize 5.536 }\sin X_{i}\\+{ 9.506+ \sum_{j=1}^{N} A_{ij} 0.002 X_{j}}
    \end{array}$ & 
   $\begin{array}{l}
        \frac{-682.033}{e^ {(-(X_{i}-5))}}+ \tiny {\sum_{j}^{N}} A_{i j}\\\tiny {({\tiny 1.499}-\frac{0.471 X_{i} X_{j}}{K_{i} n})}
    \end{array}$ & 
   $\begin{array}{l}
        \frac{-2453.529}{e^{\left(-\left(X_{i}-10\right)\right)}}+{15.621}- \\\sum_{j=1}^{N} A_{i j}(0.074 X_{j}\\+0.321 \frac{X_{j}-X_{i}}{K_{\text {in }}})
    \end{array} $& 
   $\begin{array}{l}
       \frac{-0.605 X^{5}}{1+X_{i}^{5}}+\\\sum_{j=1}^{N} A_{i j} 0.127
    \end{array}$ 
    \\ \hline
    TPSINDy$-\mathcal{H}_{R}$ &  
    $\begin{array}{l}
        -0.005 X_{i}^{3}+ \sum_{j=1}^{N} \\A_{i j}(0.859 X_{j} X_{i}+\\0.025 (X_{j}-X_{i}))
    \end{array}$ & 
   $\begin{array}{l}
        \frac{-2.154 e 11}{e^{-5(X_i-10)}}+\\\sum_{j=1}^{N} A_{i j}  0.165
    \end{array}$ & 
   $\begin{array}{l}
        \frac{-69.860}{e^{-(X_i-10)}}+ 0.725 \cos X_{i}\\+\frac{27.055 X_{i}}{K_{in} }+\sum_{j=1}^{N} A_{i j} 0
    \end{array}$ & 
   $\begin{array}{l}
        0.290- \sum_{j=1}^{N} \\A_{i j}1.078 X_{i} X_{j}
    \end{array}$ & 
   $\begin{array}{l}
     -0.451 X_{i}+0.159+ \\\sum_{j=1}^{N} A_{i j} \frac{0.013}{1+ e^{-(X_j-1)}}
   \end{array} $ &
  $\begin{array}{l}
    -0.259 X_{i}^{3}+0.138+ \\\sum_{j=1}^{N} A_{i j} \frac{-0.007\left(X_{i} X_{j}\right)^{5}}{1+\left(X_{i} X_{j}\right)^{3}}
   \end{array} $ 
    \\ \hline
    GNN+GP & 
    $\begin{array}{l}
        1.130\left(1-X_{i}\right)+\\\sum_{j=1}^{N} A_{i j}\left(X_{i} X_{j}\right)
    \end{array}$ & 
  $\begin{array}{l}
        -1.000 X_{i}+\\\sum_{j=1}^{N} A_{i j} \frac{X_{j}^{2}}{2+X_{j}^{2}}
    \end{array}$ & 
  $\begin{array}{r}
        0.722 X_{i}^{2}-0.170 X_{i}^{3}+ 1+\\\sum_{j=1}^{N} A_{i j} (0.192 X_{j} X_{i}\\+ 0.004 X_{j})
    \end{array}$ & 
    $\begin{array}{l}
  0.486 X_{i}-0.998 X_{i}^{2}\\- \sum_{j=1}^{N} A_{i, j}X_{j}X_{i}
  \end{array}$ &
  $\begin{array}{l}
        -0.596 X_{i}+\sum_{j=1}^{N} \\A_{i j} (0.237 X_{j} +0.262)
    \end{array}$ & 
  $\begin{array}{r}
        -1.000 X_{i}+\sum_{j=1}^{N} A_{i j} \\X_{j}  \left(1-X_{i}\right)
    \end{array}$ 
    \\ \hline
     LLC+TPSINDy &  
    $\begin{array}{l}
  \left(1-X_{i} \right)+\sum_{j=1}^{N} \\A_{i, j}\left(X_{i} X_{j}\right)
  \end{array}$ & 
  $\begin{array}{r}
  -2.000 X_{i}+\sum_{j=1}^{N} \\A_{i j} \frac{X_{j}^{2}}{1+X_{j}^{2}}
  \end{array}$ & 
  $\begin{array}{l}
  1.136 X_{i}^{2}-0.197 X_{i}^{3}- \\0.534 X_{i}+  \sum_{j=1}^{N} A_{i j} \\\frac{X_{i} X_{j}}{4.86+0.90X_{i}+0.10 X_{j}}
  \end{array}$ & 
    $\begin{array}{l}
  0.499 X_{i}-1.000 X_{i}^{2}- \\\sum_{j=1}^{N} A_{i j}\left(0.999\left(X_{i} X_{j}\right)\right.
  \end{array}$ & 
  $\begin{array}{r}
  -X_{i} +\sum_{j=1}^{N} A_{ij}\\ \frac{1}{1+e^{-(X_j-1)}}
  \end{array}$ & 
  $\begin{array}{l}
  -1.000 X_{i}+\sum_{j=1}^{N}\\ A_{i j} X_{j}  \left(1-X_{i}\right)
  \end{array}$ 
    \\ \hline
    LLC &  
    $\begin{array}{l}
  1-X_{i}+\sum_{j=1}^{N}\\ A_{i j}\left(X_{i} X_{j}\right)
  \end{array}$ & 
  $\begin{array}{l}
  -2.000 X_{i}+\\\sum_{j=1}^{N} A_{i, j} \frac{X_{j}^{2}}{1+X_{j}^{2}}
  \end{array}$ & 
  $\begin{array}{l}
  X_{i}^{2}\left(1.20-0.20X_{i}\right)- X_{i}+\\0.854 +\sum_{j=1}^{N} A_{i j} \\\frac{X_{i} X_{j}}{4.86+0.90 X_{i}+0.10 X_{j}}
  \end{array}$ & 
  $\begin{array}{l}
   0.500 X_{i}-1.0002 X_{i}^{2}-\\\sum_{j=1}^{N} A_{i j}X_{j} X_{i}
   \end{array}$ & 
  $\begin{array}{r}
  -X_{i}+\sum_{j=1}^{N} A_{i j}\\ \frac{e^{X_j}}{2.712+e^{X_j}}
  \end{array}$ & 
  $\begin{array}{l}
  \small{-1.000} X_{i}+\sum_{j=1}^{N} A_{i j} \\X_{j} \left(1-X_{i}\right)
  \end{array} $ \\
    \hline
\end{longtable}
\begin{tablenotes}
    \footnotesize 
    \item Note: In order to save space, we overlook the time t when presenting the equations.
\end{tablenotes}  
\end{threeparttable}
\end{landscape}

\subsection{More performance comparison}

In addition to the two indicators presented in the main text, i.e., $R^2$ score and Recall, we introduce more performance comparison metrics here to provide a more comprehensive comparison.

\begin{itemize} 
\item \textbf{Mean Relative Error.} MRE is the mean absolute difference between the predicted and true values divided by the absolute value of the true values. It considers the scale of the true value, so it is universal for data of different magnitudes, so as to provide consistent evaluation across different problems and datasets, which can be calculated as
  \[MRE = \frac{1}{TN}\sum_{i=1}^{N}\sum_{t=1}^{T}|\frac{{X}_i(t)-\hat{{X} }_i(t)}{{X}_i(t)}|, \]
where, $N$ and $T$ are the number of system nodes and the maximum prediction time, respectively. $\hat{{X} }_i(t)$ is the predicted state of node $i$ at time $t$, and ${X}_i(t)$ is the ground truth. 
\item \textbf{Mean Absolute Error.} MAE quantifies the mean absolute error between the predictions and the true values over all nodes and at all times, which can be calculated as
 \[MAE=\frac{1}{TN}\sum_{i=1}^{N}\sum_{t=1}^{T}|{X}_i(t)-\hat{{X} }_i(t)|, \]
where, $N$ and $T$ are the number of system nodes and the maximum prediction time, respectively. 
$\hat{{X} }_i(t)$ is the predicted state of node $i$ at time $t$, and ${X}_i(t)$ is the ground truth.
 \item \textbf{$\text{L}_2$ error.} This error gives an idea of how close the identified coefficients are to the true coefficients in a relative sense, which can be seen as a quantitative indicator of whether the discovered equation is correct in form.
 \[L_2~\text{error} = \frac{|\xi_{\text{pre}}-\xi_{\text{True}}|_2}{|\xi_{\text{True}}|_2},\]
 where $\xi_{\text{pre}}$ represents the equation coefficients identified by a method and $\xi_{\text{True}}$ represents true coefficients. 
 \item \textbf{Precision.} This statistic measures the percentage of correctly identified coefficients among the total number of identified coefficients, which can be calculated as
 \[P = \frac{|\xi _\text{pre}  \odot  \xi _\text{true}|_0 }{|\xi _\text{pre}|_0}, \]
 where $\odot$ represents the element-by-element product of two vectors, $|\cdot|_0$ denotes the number of non-zero elements in the vector.
\end{itemize}

\begin{figure}[h]
    \centering        
    \includegraphics[width=\textwidth]{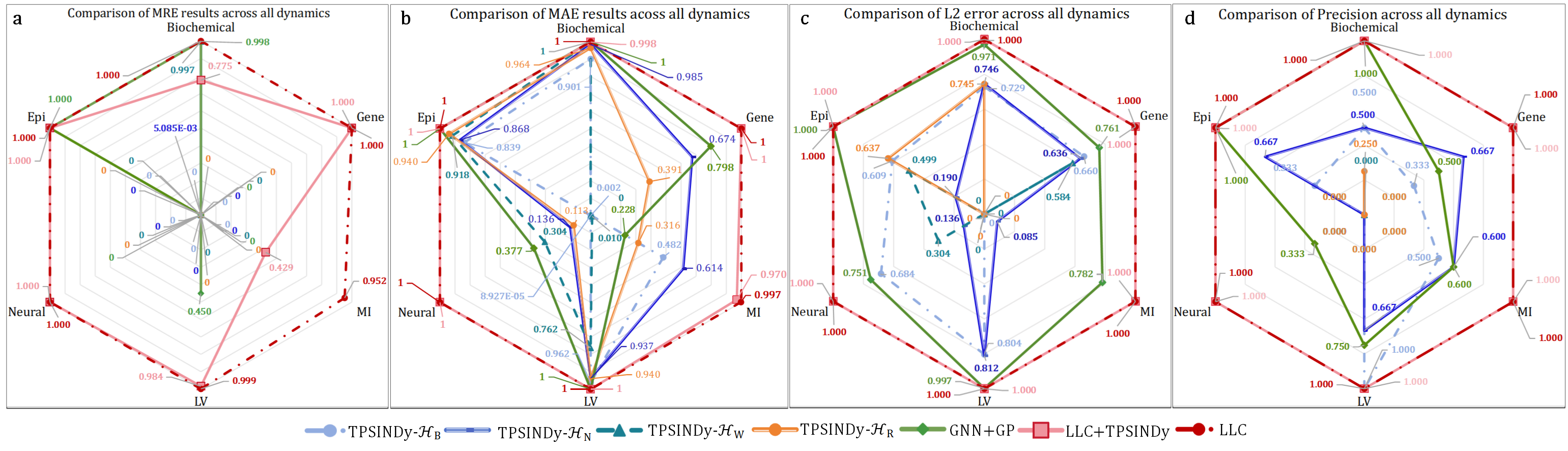}    
    \caption{
    Comparison of the accuracy on several metrics, including MRE (\textbf{a}), MAE (\textbf{b}), L$_2$ error (\textbf{c}), and Precision (\textbf{d}) for reconstructing dynamics from six scenarios, including Biochemical (Bio),  Gene regulatory (Gene), Mutualistic Interaction (MI), Lotka-Volterra  (LV), Neural (Neur), and Epidemic (Epi) dynamics.
    TPSINDy's results are highly dominated by its choice of function terms and our LLC significantly outperforms the comparative methods covering all network dynamics scenarios. 
    For the convenience of visualization, MRE, MAE, and L$_2$ error are deflated transformed and mapped into [0, 1], and then the mapped value is subtracted by 1.
    }    
    \label{figA1}
\end{figure}

Fig. \ref{figA1} shows the performance comparison on the above metrics.
MRE and MAE values closer to 0 indicate smaller discrepancies between the predicted and true values. 
Compared to the TPSINDy, the performance variations of our proposed methods on different dynamics are significantly smaller.
Notably, the results of the proposed LLC are nearly optimal. 
Furthermore, in the comparison of equation forms, the best TPSINDy version can only identify $\sim$66.7\% of the terms in the target equations, whereas our proposed LLC can identify nearly 100\% of the terms.


Apart from performance comparison, Fig.~\ref{figA2} shows comparison of results produced by various methods on the MI scenario as a visual example.
Our LLC outperforms others, which has the closest predictive curve to the true values (see Fig.~\ref{figA2}(a)).
It is evident that the TPSINDy-$\mathcal{H}_{W}$ exhibits the worst performance, followed by the redundant method. The TPSINDy-$\mathcal{H}_{N}$ approximates the true value around 200 time steps, gradually deviates from the ground truth, and eventually maintains a stable distance around 600 time steps. 
In contrast, the node trajectories generated by the proposed LLC and LCC+TPSINDy consistently match the true trajectories, thereby demonstrating the effectiveness of the proposed method compared to the TPSINDy. 
Fig.~\ref{figA2}(b) shows the designed neural network fits the differential of node activity well, and effectively decomposes signal into self and interact components, as shown in Fig.~\ref{figA2}(c-d).
To visually illustrate the discrepancy between each node’s state and the ground truth, we employ the NED heatmap, which reveals that our LLC and LCC+TPSINDy exhibit minimal differences, while others, although fitting well for some nodes, exhibit errors exceeding 1 and even reaching a maximum deviation of 3 in many other nodes, as shown in Fig.~\ref{figA2}(e).

\begin{figure}[h]
    \centering 
    \includegraphics[width=\textwidth]{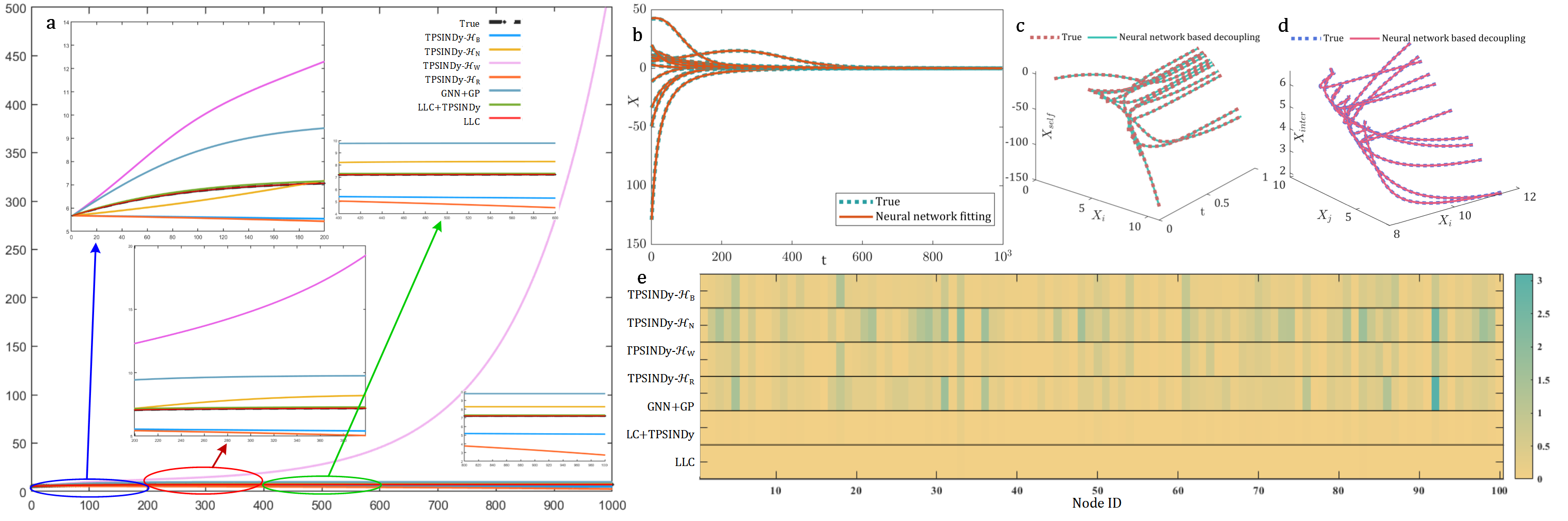}    
    \caption{
    Comparison of results produced by various methods on the MI scenario.
    \textbf{a.} Comparison of the difference between the predictive states and the true values using various methods on a node in the MI scenario.
    \textbf{b.} The fitting results of the $\dot{X}_{i}(t)$ on a node by neural networks.
    \textbf{c.} The decoupling results of the self dynamics on a node ($\Hat{Q}_{\theta_1}^{(self)}$).
    \textbf{d.} The decoupling results of the interaction dynamics on a node ($\Hat{Q}_{\theta_2}^{(inter)}$).  
    \textbf{e.} NED heat map across all nodes.} 
    \label{figA2}
\end{figure}

\section{Robustness evaluation}\label{secB}

To evaluate the robustness of our tool, we test the performance when observations are noisy or when topological structures are missing.
We choose the Kuramoto dynamics \cite{pietras2019network}, a mathematical model for studying how a set of mutually coupled oscillators can synchronize by interacting with each other, to evaluate the robustness, which can be formulated as
\begin{equation*}
    \dot{X}_i(t) = \omega_i + \epsilon   \sum_{j=1}^{N}A_{i,j}sin(X_j(t)-X_i(t)),
\end{equation*}
where $X_j(t)$ denotes the phase of the $i$-th oscillator, $\epsilon$ = 0.015 is the coupling strength, and $\omega_i$ is the natural frequency distributed according to a given normal distribution $\mathcal{N}(1,\sigma)$ with $\sigma = 1$. 
{The time step was set to $\delta t = 0.01$, $T = 5$ and $T_{end}=100$.}
We assume that the underlying topological structure is a ER network with $N=100$.

Complex network systems typically have two sources of noise: measurement and topological noises.
The former is mainly caused by inaccurate measurement results from unstable sensors.
The latter is due to the deletion or addition of nodes or edges, resulting in incomplete captured topology \cite{sase2016estimating}. 
Herein, we use the Add White Gaussian Noise (AWGN) algorithm to generate noisy observation data in the MatLab environment.
The specific noise addition operation is as follows:
\begin{equation*}
\begin{array}{c}
{X}_i^{noise}(t) = X_i(t)+\beta \times \mathcal{N}(0,1) X_i(t),\\  
\dot{X}_i(t) = Q^{self}(X_i(t)) + \sum_{j=1}^{N}A_{i,j}^{'}Q^{inter}(X_i,X_j), \\
\end{array}  
\end{equation*}
where, $\beta$ is the noise intensity,
$ A_{i,j}^{'} = A_{i,j} \cdot (1 - \mathbb{I}{\{A_{i,j} = 1 \text{ and } R < \eta\}}) + (1 - A_{i,j}) \cdot \mathbb{I}{\{A_{i,j} = 0 \text{ and } R < \eta\}}$.
A random variable $R$ is drawn from a uniform distribution within the range [0, 1]. 
The spurious links probability is designated as $\eta$. 
$\mathbb{I}$ is an indicator function that assumes the value of 1 when a condition is met and 0 otherwise.
The amount of noise can be measured by the signal-to-noise ratio (SNR) \cite{johnson2006signal}, and the larger its value, the less noise is added.

\begin{figure}[h]
    \centering
    \includegraphics[width = 0.8\textwidth]{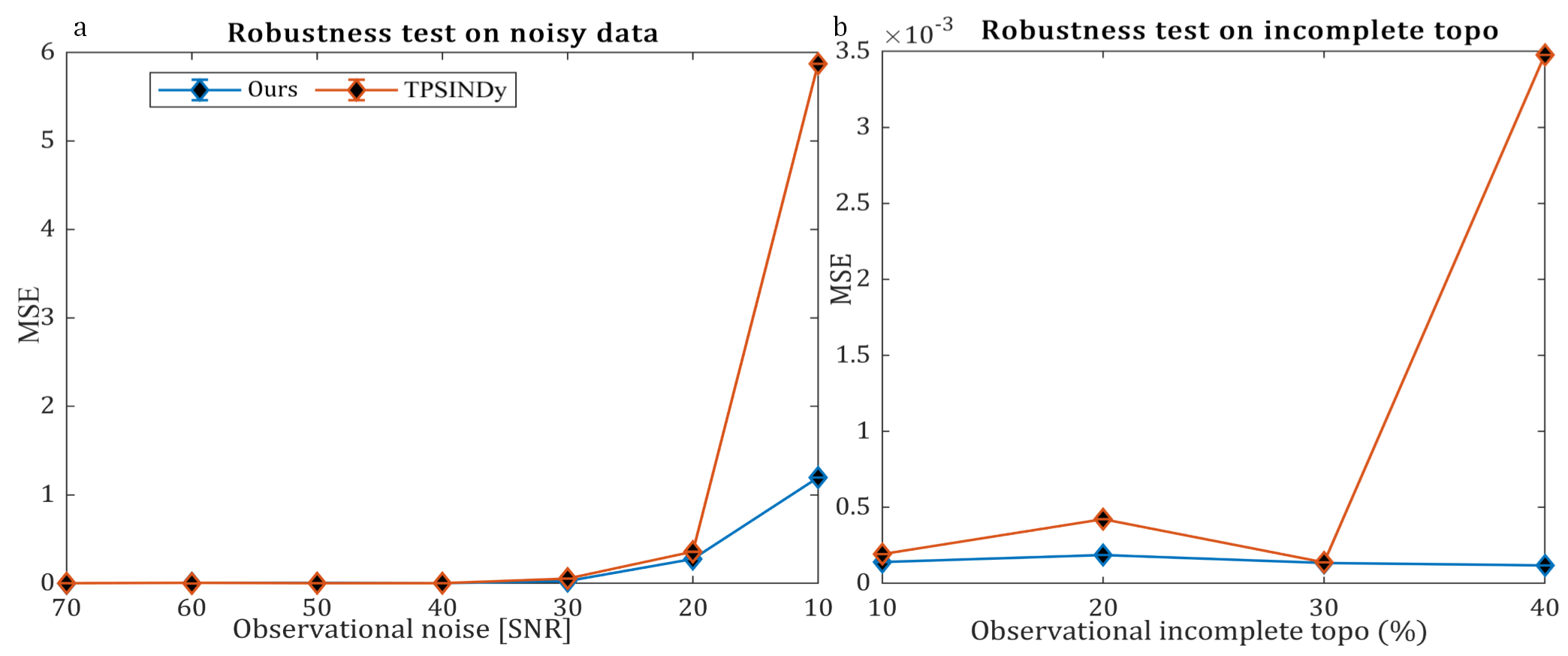}    
    \caption{Robustness evaluation. \textbf{a.} Comparison of performance (MSE) with increasing observation noise. \textbf{b.} Comparison of performance (MSE) as the percentage of spurious topology increases.}
    \label{figc1}
\end{figure}

Fig.~\ref{figc1}(a) shows the comparison of performance (MSE) with increasing observation noise.
As the SNR drops from 70 dB to 30 dB, our tool maintains an MSE close to 0,
while more noises weaken performance but still significantly outperforms the TPSINDy, which keeps increasing exponentially. 
Fig.~\ref{figc1}(b) shows the comparison of performance (MSE) as the percentage of missing topology increases, where the horizontal axis represents the percentage of missing edges.
The results show that both methods have the extension completion ability, but the performance of our tool can accommodate more missing information.

\section{Multi-Dimensional and Heterogeneous Network
Dynamics}\label{secC}

\subsection{FitzHugh-Nagumo dynamics}

The FitzHugh-Nagumo model (FHN) \cite{fitzhugh1961impulses} is a neuron model used to describe the excitatory behavior of neurons, which is a simplified version of the Hodgkin-Huxley model and mainly used to study the action potentials of neurons. 
It can be defined by a set of two ordinary differential equations that capture the main features of excitability in neural membrane dynamics. 
The equations controlling the dynamics are as
\begin{equation*}
    \label{eqfhn}
    \left\{
    \begin{array}{l}
         \dot{X}_{i,1}(t) = X_{i,1}(t)-X_{i,1}(t)^{3}-X_{i,2}(t)-\epsilon \sum_{j=1}^{N}A_{i,j}\frac{X_{i,1}(t)-X_{j,1}(t)}{K_{in}} ,\\
         \dot{X}_{i,2}(t) = a+bX_{i,1}(t)+cX_{i,2}(t) 
    \end{array}
    \right.
\end{equation*}
where the first component $X_{i,1}$ denotes the membrane potential containing the self and interaction dynamics, $K_{in}$ is the in-degree of neuron $i$ (denoting the number of afferent connections to node $i$), and $\epsilon = 1$. The second component $X_{i,2}$ denotes the recovery variable, where $a = 0.28$, $b = 0.5$, $c = - 0.04$. 
We set the underlying topology to BA network with $N=100$ and terminal time $T = 3$, $T_{end} = 100$ and time step $\delta t = 0.01$. 

Moreover, we apply our LLC to FHN on a Barabási-Albert network and two empirical networks, including the C. elegans \cite{yan2017network} and Drosophila \cite{scheffer2020connectome}, to analyze whether different topologies have an impact on the results.
The experimental setup is delineated as follows. 
During the training phase, we conduct experiments across three network topologies: BA, C.elegans, and Drosophila. 
The initial value is set to $N(0,1)$, with a time interval of 0.01 and a total duration of T = 10. 
We systematically identify the equations for each topology before performing comparative analyses on a specified network structure (e.g., BA) to evaluate experimental trajectories and node errors. 
The detailed results are illustrated in Fig. \ref{figfhn}(a). 
It is evident that our tool effectively identifies the correct equations in both synthetic and real-world structures. 
The long-term predictive trajectories produced by the discovered equations derived from different topologies exhibit remarkable consistency. 
Fig. \ref{figfhn}(b) visualizes node error distributions; notably, approximately 90\% of nodes demonstrate NED error close to zero, with only a few exhibiting errors around 0.1. 
The relative errors associated with the derived equation coefficients are presented in Fig. \ref{figfhn}(c), where maximum coefficient discrepancies do not exceed 3\%. 
Finally, spatial trajectory comparisons depicted in Fig. \ref{figfhn}(d) indicate that LLC demonstrates robust fitting capabilities across all dimensions.
\begin{figure}[h]
    \centering
    \includegraphics[width=\textwidth]{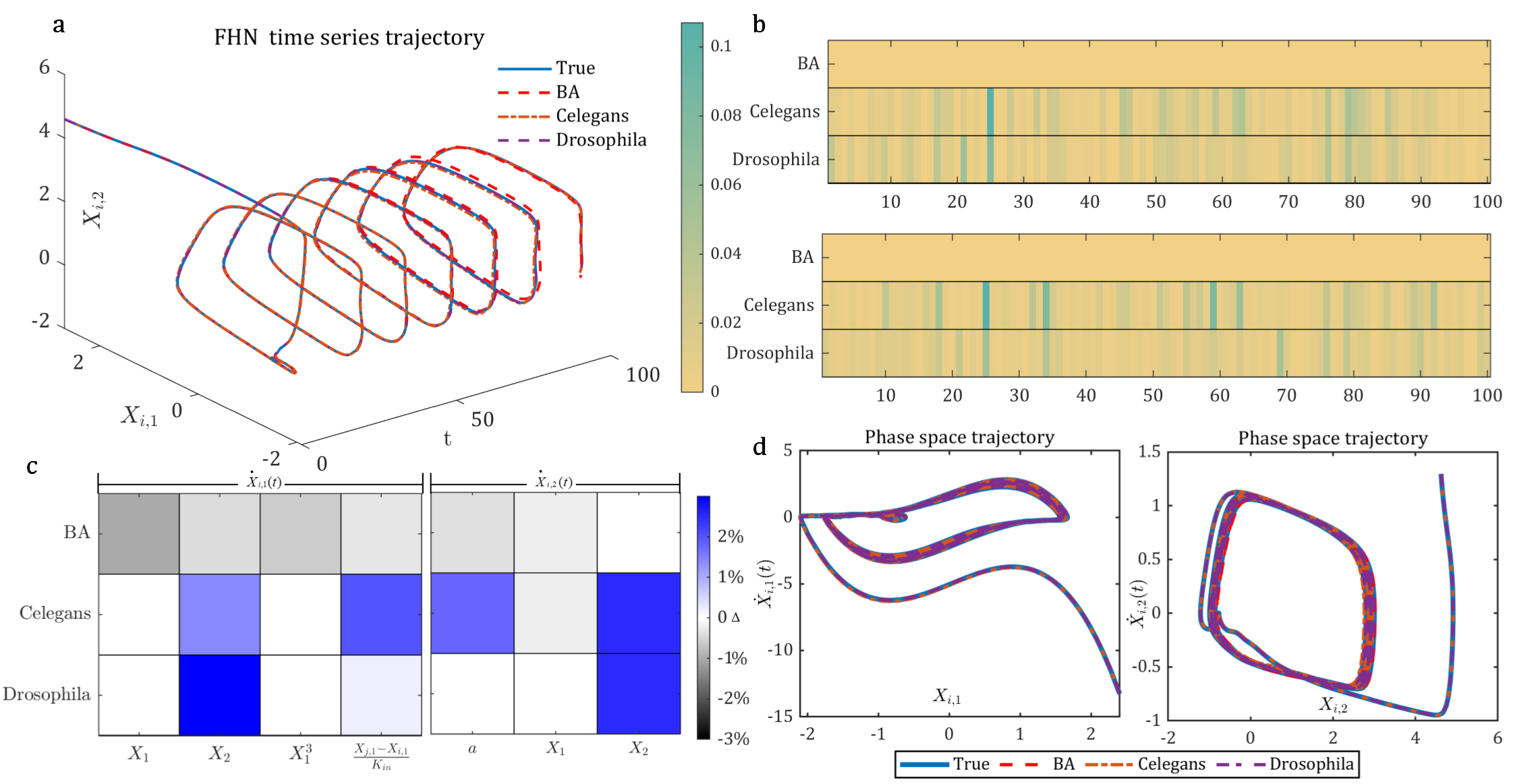}    
    \caption{Recognition results on different topological structures. \textbf{a.} Comparison of node activities for discovered equations. 
    \textbf{b.} NED values of nodes.
    \textbf{c.} Comparison of coefficient errors for discovered equations. 
    \textbf{d.} Phase space trajectories of a specified node. 
    The results empirically demonstrate that our tool can obtain satisfactory equations and predictive trajectories on both synthetic and empirical topologies.
    }
    \label{figfhn}
\end{figure}

\subsection{Predator-prey system}
The predator-prey (PP) is a heterogeneous system \cite{chen2014minimal}, where node state is the individual's position, but nodes are divided into two roles, i.e., {a predator ($i=0$) and many preys ($i>0$)}, resulting in three types of pair-wise relationships, i.e., predator-prey, prey-predator, and prey-prey.
The interactions between prey can be modeled as exhibiting paired short-range repulsion and long-range attraction. Specifically, we consider the form of the prey-prey interaction as follows:
\[F_{i,\text{prey-prey}} = \frac{1}{N} \sum_{j=1, j \neq i}^{N} \left( \frac{1}{|X_i(t) - X_j(t)|^2} - a \right) (X_i(t) - X_j(t)).\]
Here, $\frac{X_i(t) - X_j(t)}{|X_i(t) - X_j(t)|^2}$ represents a Newtonian short-range repulsive force directed from \(X_i(t)\) to \(X_j(t)\), while \(-a (X_i(t) - X_j(t))\) denotes a linear long-range attractive force in the same direction. Although a more generalized attraction-repulsion dynamic can be considered, we focus on this specific form as it allows for clearer and more definitive results. 
Assuming the presence of a predator, we set its position to be denoted by \(X_0(t)\).
We consider the predator's interaction with the prey as a repulsive force, expressed as: \[F_{j,\text{predator-prey}} = b \frac{X_j(t)-X_0(t)}{|X_j(t) - X_0(t)|^2},\]
where, \(b\) represents the intensity of the repulsive force. 
Based on the above considerations, the predator-prey system can be formulated as follows:
\begin{equation*}
\left\{
    \begin{array}{l}
        \dot{X}_0(t)=\frac{c}{N} \sum_{j=1}^{N}\frac{X_{j}(t)-{X}_0(t)}{\left|X_{j}(t)-{X}_0(t)\right|^{2}}, \\
        \dot{X}_{i}(t)=b\frac{\left(X_{i}(t)-{X}_0(t)\right)}{\left|X_{i}(t)-{X}_0(t)\right|^{2}}+ \frac{1}{N}\sum_{j=1}^{N} \left(\frac{(X_{i}(t)-X_{j}(t))}{\left|X_{i}(t)-X_{j}(t)\right|^{2}}+a\left(X_{j}(t)-X_{i}(t)\right)\right),i>0,
    \end{array}
    \right.
\end{equation*}
where we set $a$, $b$, and $c$ to 1.0, 0.2, and 0.7, respectively. 
\begin{figure}[h]
    \centering 
    \includegraphics[width=\textwidth]{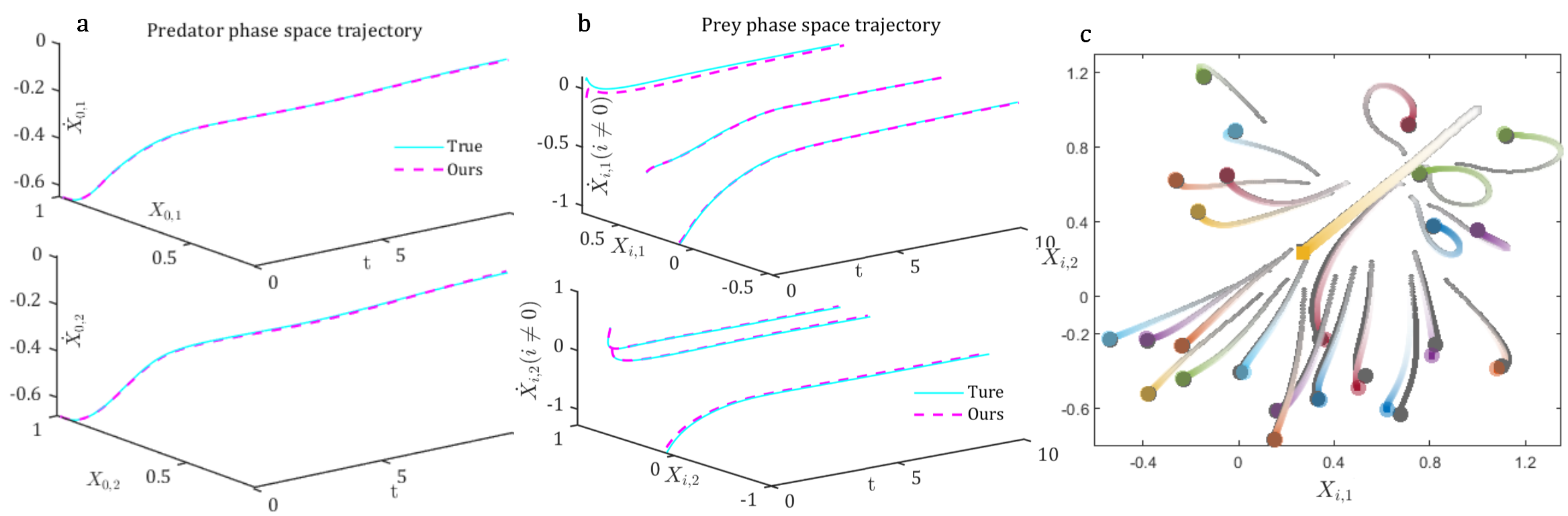}   
    \caption{
    Comparison of trajectories of the Predator-Prey system. 
    \textbf{a.} Comparison of phase trajectories for each dimension of the predator. 
    \textbf{b.} Comparison of phase trajectories for each dimension of a prey. 
    \textbf{c.} Comparison of trajectories of predator and prey swarm, where gray trajectories represent the ones generated by the true equation and color ones are generated by the equation discovered by our LLC. Square represents predator and circles represent prey.}
    \label{figpp}
\end{figure}
For the underlying topology, we consider a fully connected structure in this system with $N=101$.

Fig.~\ref{figpp} shows the comparison of trajectories of the predator-prey system.
From Fig. \ref{figpp}(a, b), we see that the proposed LLC fits the real spatial trajectories very well.
Additionally, we show how well the inferred equations fit the predicted and actual trajectories. 
Fig. \ref{figpp}(c) provides an example with 50 agents, where the colored part represents the predicted trajectories and the gray part represents the actual trajectories. 
Both coincide to a large extent, indicating that the equation discovered by LLC predicts this system well. 

\section{Chaotic Networks Dynamics}\label{secD}
\subsection{Lorzen system}
Coupled Lorzen system \cite{MOGHTADAEI2012733} exhibits chaotic properties, in particular sensitive dependence on initial conditions, which means that even small changes in initial conditions can lead to significantly different system behavior. 
To verify whether the proposed LLC is sensitive to initial values, we apply it to the coupled Lorzen system systems with different initial values and examine whether the discovered governing equations are consistent.

Specifically, we apply our LLC to a coupled Lorzen system \cite{MOGHTADAEI2012733} governed by
 \begin{equation*}
 \label{eqLorenz}
    \left\{\begin{array}{l}
       \dot{X}_{i,1}(t)=a(X_{i,2}(t)-X_{i,1}(t))+\epsilon {\textstyle \sum_{j=1}^{N}} A_{i,j}(X_{j,1}(t)-X_{i,1}(t)),\\
       \dot{X}_{i,2}(t)=rX_{i,1}(t)-X_{i,1}(t)X_{i,3}(t)-X_{i,2}(t),\\
       \dot{X}_{i,3}(t)=X_{i,2}(t)X_{i,1}(t)-bX_{i,3}(t),
    \end{array}\right.
\end{equation*}
where $a=10$, $\epsilon=0.05$, $r=28$ and $b=\frac{10}{3}$ are system parameters. 
Here, the interactions are assumed to occur between the first component $X_{i,1}$ without lack of generalization. 
Then, we employ a BA network with $N=100$ as the topological structure to produce the dynamics data of the Lorenz.
We set the end time $T = 3$, {$T_{end}=100$} and time step $\delta t = 0.01$.

We establish three distinct initial conditions as follows: 
\begin{itemize}
    \item Initial condition 1: A Gaussian distribution with zero mean and one variance; 
    \item Initial condition 2: A uniform distribution within the range (0, 2);
    \item Initial condition 3: A fixed initial value of 0.1.
\end{itemize}

In addition to the consistent equations discovered at three different initial values presented in the main text, we also demonstrate the inferring results of a coupled Lorzen system with 2,500 attractors under different initial conditions, as shown in Fig.~\ref{figLorenz}.
It is evident that despite only minor variations among the initial conditions, there are pronounced differences in the final states, as demonstrated by comparing the first two rows.
Notably, our LLC yields results akin to those presented in the second row regardless of varying initial conditions, as shown by comparing the second and third rows.
To elucidate the distinctions between ground truth and predictive outcomes more clearly, we visualize the error between them at final state, indicating that our LLC can achieve predictions with an error margin approaching zero.

\begin{figure}[h]
    \centering
    \includegraphics[width=\textwidth]{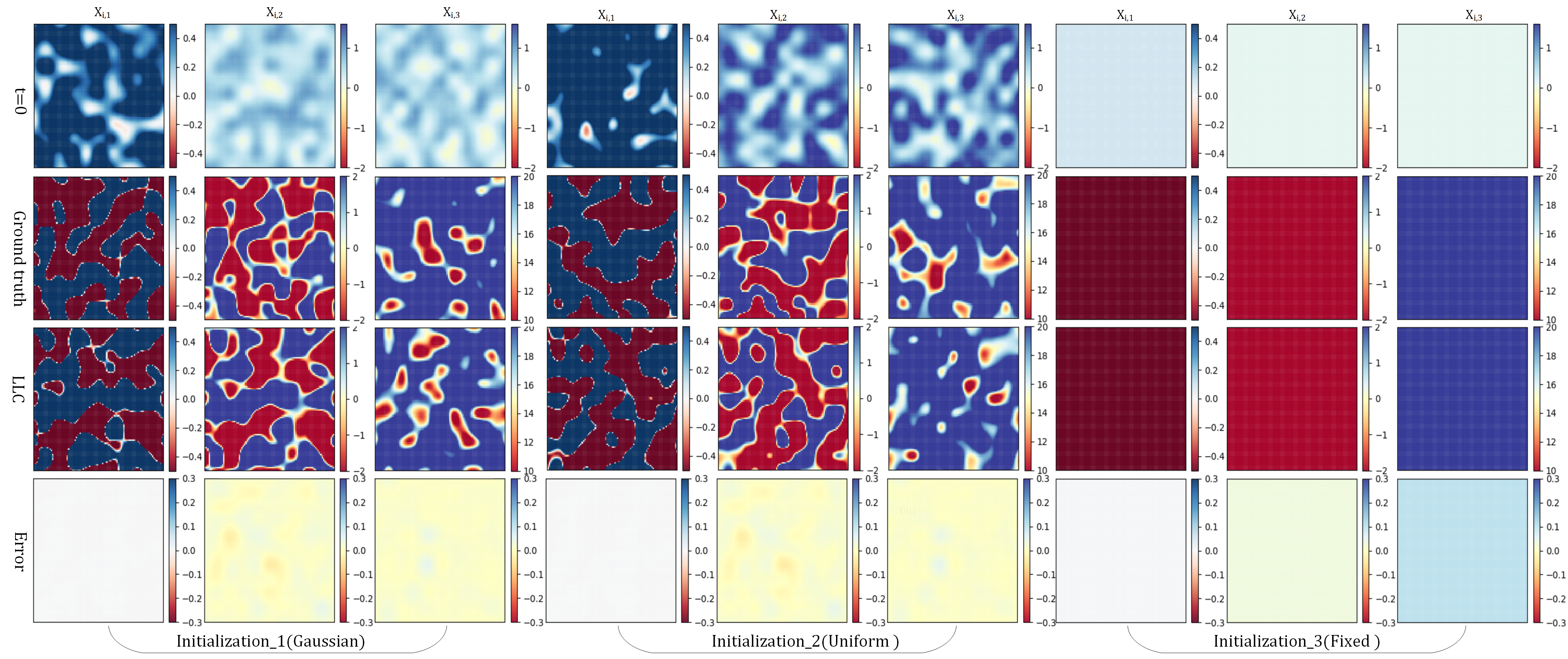}    
    \caption{Inferring results of a coupled Lorzen system with 2,500 attractors under different initial conditions.
    We rearranged all attractors in a $50\times 50$ grid format.
    The first row shows the initial state of each dimension, i.e., $t=0$. 
    The second row shows the final state ($T_{end}=30$) of the ground truth equation under different initial conditions.
    The third row shows the predictive final state ($T_{end}=30$) produced by equations discovered by the LLC under different initial conditions. 
    The last row shows the error between the predictive final state and the ground truth.}
    \label{figLorenz}
\end{figure}

\subsection{Rössler system}

By comparing the TPSINDy on the Rössler system, our LLC restores more accurate governing equations (Fig.~\ref{fig5}(c)) and has smaller predictive errors (Fig.~\ref{fig5}(d)).
We also analyze the state transition behavior of the Rössler system, and it is evident from the comparison of bifurcation diagrams generated by the inferred and true equations that period-doubling patterns and the bifurcation points are consistent, showing the transition process from period-1 to period-2, to period-4, to chaos, then to period-3, and finally to chaos again (see Fig.~\ref{fig5}(e)).

We also consider a coupled Rössler system, which are often used in classical models to study chaotic dynamics and synchronization phenomena in complex systems. 
The chaotic states data is generated by the governing equations as
\begin{equation}
\label{eqRoss}
    \left\{\begin{array}{l}
\dot{X}_{i,1}(t)=- X_{i,2}-X_{i,3}+\epsilon {\textstyle \sum_{j=1}^{N}} A_{i,j}(X_{j,1}-X_{i,1})\\
\dot{X}_{i,2}(t)=  X_{i,1}+aX_{i,2}\\
\dot{X}_{i,3}(t)=b+X_{i,3}(X_{i,1}-c)  
\end{array}\right.
\end{equation}
where $\epsilon=0.15$, $a=0.2$, $b=0.2$ and $c=5.7$ are system parameters.
We employ a BA network with $N=100$ as the topological structure to produce the dynamics data of the system.
We set the end time $T = 3$, {$T_{end}=100$} and time step $\delta t = 0.01$.

Fig. \ref{fige7} shows the comparison of inferred results on a coupled Rössler system. 
LLC and LLC+TPSINDy can closely fit the original trajectories, while TPSINDy only achieves a good fit during the initial time segment, but differ significantly from the true state after exceeding ~3,000 steps, as shown in Fig. \ref{fige7}(a).
\ref{fige7}(b) shows the comparison of coefficients of the equations discovered by different methods.
We see that the coefficients of the equation discovered by TPSINDy slightly differ from the true coefficients.
This also shows that, for chaotic systems, although only a small coefficient change, it will also have a large impact on the predictive states, especially for long-term forecasting.
\ref{fige7}(c) shows the NED values of each node under different methods, demonstrating that our LLC has a relatively high accuracy.
For the first two dimensions, the TPSINDy consistently produces NED values greater than 0.1, while the NED values of our LLC mainly remain below 0.06. 
In the third dimension, the performance of the proposed LLC is worse than the TPSINDy method. 
However, the combination of LLC+TPSINDy effectively addresses this shortcoming, resulting in NED values close to 0.

 \begin{figure}[h]
    \centering 
    \includegraphics[width=1.0\textwidth]{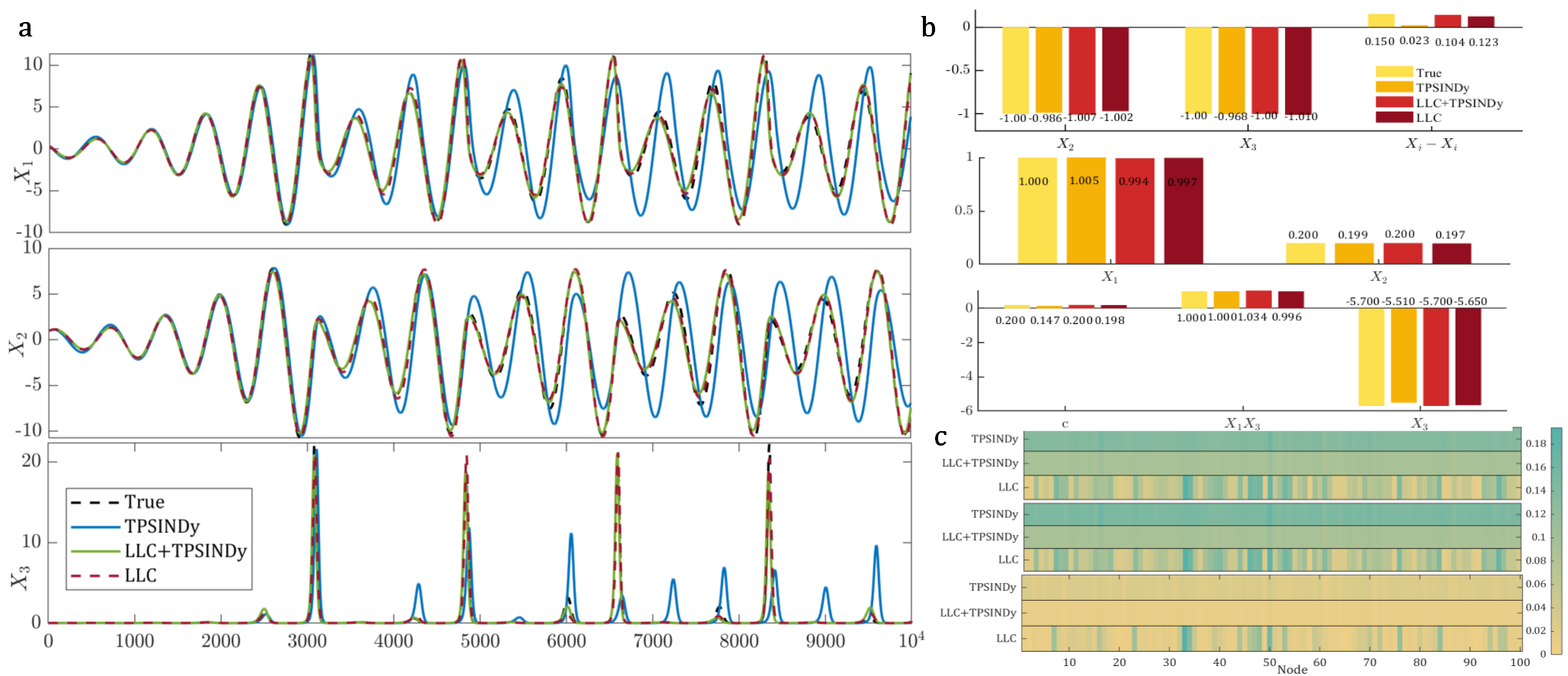}    
    \caption{Comparison of inferred results on a coupled Rössler system. 
    \textbf{a.} Comparison of predictive states from different methods in each dimension changes over time.
    \textbf{b.} Comparison of coefficients of the equations discovered by different methods.
    \textbf{c.} NED values of each node under different methods.}  
    \label{fige7}
\end{figure}

To more intuitively demonstrate the chaotic nature of the system, we utilize a Poincaré surface to construct a bifurcation diagram, as shown in Fig.~\ref{fig5}(e) in the main text. 
Specifically, we select \( X_1 = 0.1 \) as the section, recording intersection points as the trajectory crosses this plane. 
By documenting a substantial number of intersection points, we can generate the Poincaré section.
If the system exhibits periodic motion, a finite number of points will appear on the Poincaré section. 
For instance, period-2 will manifest as two points on the section. 
Conversely, if the system exhibits chaotic motion, the Poincaré section will display a large number of densely packed points, revealing a complex structure.
Fig.~\ref{fige8} visualizes the period-doubling and chaos of the Rössler system,
showing the consistent limit cycles and chaotic phenomena.

 \begin{figure}[h]
    \centering 
    \includegraphics[width=0.8\textwidth]{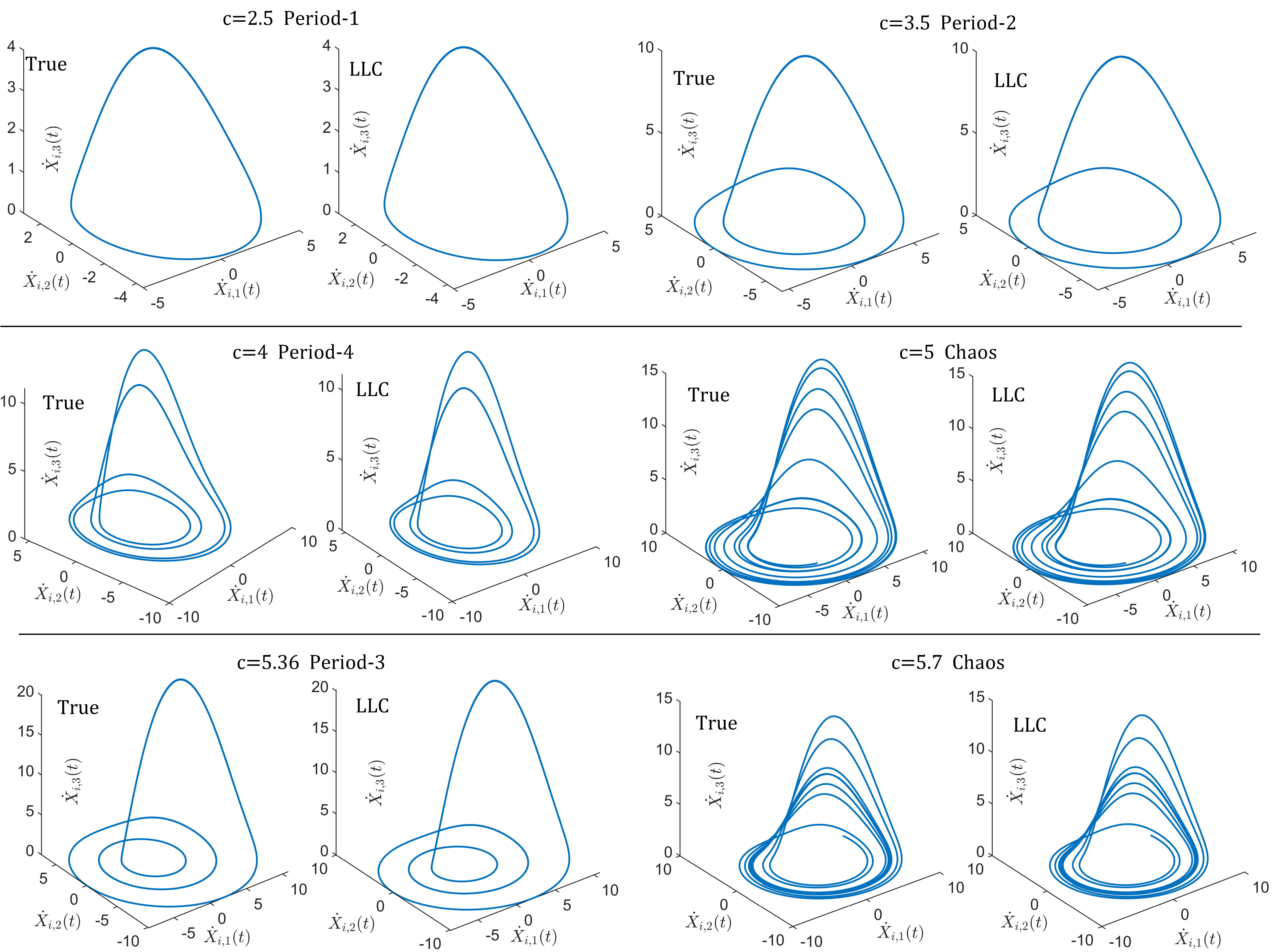}    
    \caption{Period-doubling and chaos of the Rössler system.
    Comparison of limit cycles at period-1 with $c=2.5$, period-2 with $c=3.5$, period-4 with $c=4$, chaos with $c=5$, period-3 with $c=5.36$, and chaos again with $c=5.7$.
    }
    \label{fige8}
\end{figure}

\section{Empirical Systems}\label{secE}
For the real-world systems, we use two datasets from different domains, including the daily global propagation data of real diseases \cite{dong2020interactive} and real-world crowd trajectory dataset \cite{boltes2013collecting}.

\subsection{Real-world global epidemic transmission}
We collect daily global spreading data on COVID-19 \cite{dong2020interactive}, and use the worldwide airline network retrieved from OpenFights \cite{openflights} as a directed and weighted empirical topology to build an empirical system of real-world global epidemic transmission.
Only early data before government intervention, i.e., the first 45 days, are considered here to maintain the spread characteristics of the disease itself.
For example, if a country reports its first case on January 19, data from January 19 to March 3 are used. It is worth noting that although the start of transmission varies across countries, in the interaction dynamics $Q_{i,j}^{(inter)}$, time $t$ corresponds to the same calendar date for all nodes. 
Our setup is consistent with \cite{gao2022autonomous}.
Fig.~\ref{figf9} shows the comparative results of the number of cases over time in sufficient countries or regions generated by TPSINDy, LLC, and LLC+TPSINDy, demonstrating the effectiveness of our LLC in discovering new symbolic models for real scenario with unknown dynamics.

\begin{figure}[h]
    \centering
    \includegraphics[width=\textwidth]{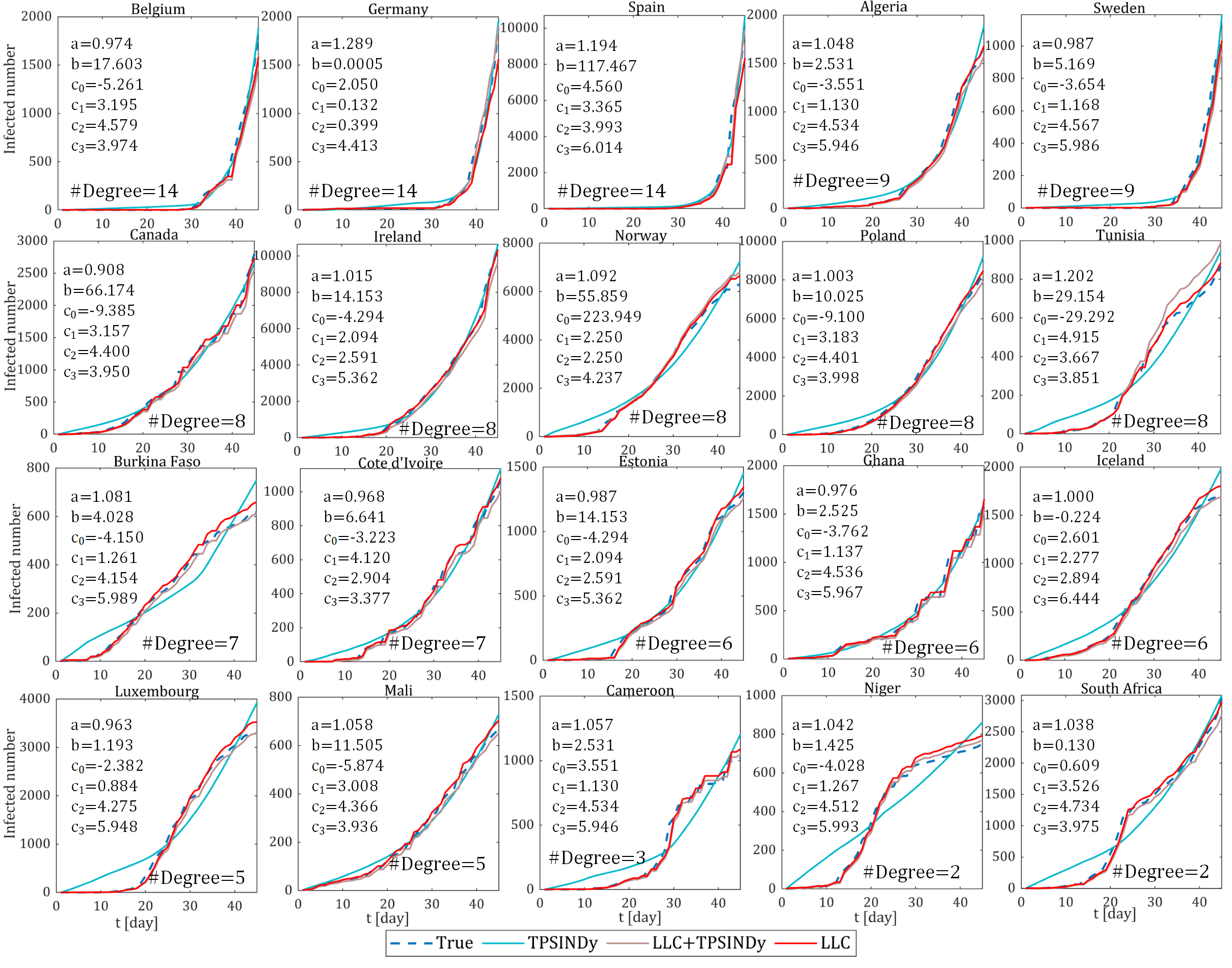}    
    \caption{Comparative results of the number of cases over time in sufficient countries or regions generated by TPSINDy, LLC, and LLC+TPSINDy.
    }
    \label{figf9}
\end{figure}

\subsection{Pedestrian dynamics}
For the crowd trajectory dataset, we use an experimental dataset from a pedestrian dynamics study: unidirectional flow in a corridor: a group of people crossing a corridor in the same direction \cite{boltes2013collecting}. Specifically, the experimental data collected at a fixed time interval of 0.04 seconds and established the training duration $T$ as 150. 
The center of the corridor was designated as the coordinate origin, resulting in a range for X of $[-5, 5]$, where -5 denotes the starting point and 5 indicates the destination. 
In this work, 17 participants commenced their journey from the initial point towards the destination.
During the testing phase, we selected trajectories from all participants over the last 30 time steps to serve as testing set.

\end{appendices}
\end{document}